\documentclass[review]{elsarticle}

\usepackage{lineno,hyperref}
\usepackage{graphicx}
\usepackage{subfigure}  
\usepackage{textcomp}
\usepackage{amsmath}
\usepackage{color}
\usepackage{array, multirow, bigdelim, makecell, booktabs} 

\usepackage{mathrsfs}
\modulolinenumbers[5]

\journal{Journal of Image and Vision Computing}

\begin{document}
	
	\begin{frontmatter}
		
		\title{Dense Graph Convolutional Neural Networks on 3D Meshes for 3D Object Segmentation and 
			Classification}
		
		\author[mymainaddress,mysecondaryaddress,mythirdaddress]{Wenming~Tang}
		\author[mymainaddress,mysecondaryaddress,mythirdaddress,myfourthaddress]{Guoping~Qiu\corref{mycorrespondingauthor}}
		%
		\cortext[mycorrespondingauthor]{Corresponding author}
		\ead{guoping.qiu@nottingham.ac.uk (G. Qiu).}
		
		\address[mymainaddress]{College of Electronics and Information Engineering, Shenzhen University, Shenzhen, China}
		\address[mysecondaryaddress]{Guangdong Key Laboratory of Intelligent Information Processing, Shenzhen University, Shenzhen, China}
		\address[mythirdaddress]{Shenzhen Institute of Artificial Intelligence and Robotics for Society, Shenzhen, China}
		\address[myfourthaddress]{School of Computer Science, The University of Nottingham, UK}
		
		\begin{abstract}
			This paper presents new designs of graph convolutional neural networks (GCNs) on 3D meshes for 3D object segmentation and classification. We use the faces of the mesh as basic processing units and represent a 3D mesh as a graph where each node corresponds to a face. To enhance the descriptive power of the graph, we introduce a 1-ring face neighbourhood structure to derive novel multi-dimensional spatial and structure features to represent the graph nodes. Based on this new graph representation, we then design a densely connected graph convolutional block which aggregates local and regional features as the key construction component to build effective and efficient practical GCN models for 3D object classification and segmentation. We will present experimental results to show that our new technique outperforms state of the art where our models are shown to have the smallest number of parameters and consietently achieve the highest accuracies across a number of benchmark datasets. We will also present ablation studies to demonstrate the soundness of our design principles and the effectiveness of our practical models.
		\end{abstract}
		
		
	\end{frontmatter}
	
	
	\section{Introduction}
	The increase in computing power and the emergence of various high-efficiency deep learning architectures have rapidly advanced many areas of computer vision. However, most of the deep learning architectures are designed for 1D (Language) and 2D (Image) data. The main reason is that deep learning requires a large amount of training data while the cost of acquiring 3D data is much higher than that of 1D and 2D, and the representation of 3D data is much more complicated. Therefore, the application of deep learning in 3D is not as efficient in 2D~\cite{ahmed2018survey}. Fortunately, with the development of 3D sensor technology, the cost of 3D data acquisition is getting lower and lower, which makes the application of deep learning increasingly popular in the 3D computer vision community.
	\par In 3D computer vision, 3D mesh is a very effective form for representing 3D objects and occupies an important position in 3D computer vision and computer graphics. Relying on the combination of geometrical structures of vertices, edges and faces, 3D mesh has the advantages of containing rich details, representing specific geometric features, and occupying a small memory footprint. Therefore, 3D mesh has obvious advantages over point clouds and voxels in describing 3D shapes. Among the many types of 3D mesh, 3D watertight mesh is the most common one. Watertight means that the mesh on all of the surfaces is complete, the lines of the mesh create valid elements, and the mesh connects to adjacent surfaces around the perimeter so that the volume is fully enclosed. Its geometric characteristics have not only become a research hotspot, but also have been widely used in games, 3D animation, 3D printing and scanning, industrial manufacturing, etc. 
	\par Classification and Segmentation are the two fundamental tasks in computer vision. Deep learning has achieved great success in 2D image processing, such as classification and segmentation~\cite{deng2009imagenet,lin2014microsoft}. However, in 3D computer vision, the irregularities of the data and the complexity of the 3D geometric structure make the classification and segmentation tasks more challenging. Nevertheless, researchers have attempted various methods, such as converting 3D data into image data. For example, by using multi-view images to express 3D shapes~\cite{su2015multi}, thereby obtaining data that can be directly input to the Convolutional Neural Networks(CNN), or by designing a unique multi-layer perception network (PointNet) to implement 3D point cloud classification and segmentation tasks~\cite{qi2017pointnet}. PointNet is the first work where the original 3D data is directly input into the neural network without conversion. Since PointNet only relies on 1D convolution of 3D point cloud, it is difficult to capture the point cloud neighborhood information. In a subsequent work, the authors proposed an improved version of PointNet (PointNet++~\cite{qi2017pointnet++}). The addition of point cloud neighborhood information improves the multi-layer perceptron network's ability to perceive the local area of the point cloud, which enhanced the network's classification and segmentation capabilities. Continuing from this technical route, Jiang et al.~\cite{jiang2018pointsift} proposed a new network named PointSIFT. They brought SIFT to 3D point clouds which improved the performance of the network in 3D point cloud learning. The neighborhood information of 3D data is not fixed and unique as in image, so the results obtained by different neighborhood constructions can vary greatly. Therefore, when applying the deep learning framework to 3D data, enhancing the perception of local (neighborhood) information is an effective method to improve network performance.
	\par Meanwhile, deep learning on 3D mesh has made great progress, and some excellent work has appeared the literature~\cite{feng2018meshnet,kostrikov2018surface,kumawat2019lp,hanocka2019meshcnn}. Yutong et al.~\cite{feng2018meshnet} designed a MeshNet network by using mesh data. The MeshNet takes the face of the mesh as a unit and uses the index of the face's three neighborhood faces and its own geometric characteristics to achieve mesh classification and retrieval. MeshNet only uses the index of the neighborhood face, but other geometric features of the neighborhood face deserve further study. Hanocka et al.~\cite{hanocka2019meshcnn} proposed a CNN network based on the edge of the mesh and named it MeshCNN. Based on the geometric characteristics of two faces on each edge, the pooling of the mesh was designed through the collapse of the edge. The unique CNN network achieved good performances in 3D mesh classification and segmentation. MeshCNN feeds 3D mesh data in the non-Euclidean space to a CNN network implemented by 2D convolution.
	\par However, it's well known that a 3D mesh is a graph with a natural topology composed of vertices and edges. Therefore, we can directly utilize its topological structure and geometric characteristics without having to convert the graph structure to adapt to a deep learning framework. In this way, we can use the original geometric characteristics of the 3D mesh, which is more intuitive and general. 3D mesh’s geometric topology and non-Euclidean spatial characteristics make it well suited for graph convolutional neural networks (GCNs)~\cite{Milano20NeurIPS-PDMeshNet}. GCNs are designed to solve non-Euclidean spatial data problems and realise the conversion between the spatial domain and the spectral domain through the Laplacian operator, which establishes associations between graph nodes. Better exploiting the relations between the graph nodes is one of the methods that researchers can use to improve the performance of deep learning in 3D data.
	\par In this paper, we present densely connected graph convolutional networks on 3D meshes (MDC-GCNs). Aiming to make 3D mesh data as convenient as image data to enter the deep learning framework, we convert 3D meshes into graph structure data. We also introduce a novel set of features for each node to enhance the expressive power of the nodes. The new features include the space coordinates of the vertex $P$, the normal of the vertex $N_v$, the Gaussian curvature of the vertex $GC$, the normal of the face $N_f$, and the angle between the faces $\theta$. MDC-GCN is a deep architecture that utilises the densely connected mechanisms of the GCN to integrate local and non-local features. Hence, this network has better learning capabilities for 3D mesh than the normal GCNs. The main contributions of our paper are as follows:
	\begin{itemize}
		\item A novel graph data structure for representing 3D mesh data. We first convert the 3D mesh to a graph where each face on the mesh has a corresponding node on the graph. A novel 1-ring neighbourhood structure of the face (see Fig.~\ref{fig:GCN_node}) is constructed to derive a multi-dimensional feature vector to represent the node. Each node’s feature vector includes the geometric  characteristics of the face and descriptions of its physical connection with its 1-ring neighbours.
		\item  A novel graph convolutuional neural network  architecture on 3D meshes for achieving state of the art results in 3D object segmentation and classification. Using the new graph representation of 3D mesh, we construct densely connected graph convolutional neural networks for implementating 3D object segmentation and classification. We present extensive experimental results to show that our new method outperformed state of the art.
	\end{itemize}
	\section{Related work}
	\par In this section, we will briefly review three related topics: densely connected convolutional networks, graph convolutional networks, and graph convolutional networks application in 3D mesh.
	\subsection{Densely Connected Convolutional Networks}
	\par In recent two decades, deep learning has played a pivotal role in computer vision. In different applications, researchers have designed different networks. As the complexity of the task increases, the network design becomes deeper and deeper, which will bring about the vanishing gradient problem. To solve this issue, researchers have proposed many solutions, such as: ResNets~\cite{he2016deep}, Highway Networks~\cite{srivastava2015training}, Fractal Nets~\cite{larsson2016fractalnet}, and Stochastic depth networks~\cite{huang2016deep}. Gao et al.~\cite{huang2017densely} designed a dense connection network (DenseNets) in which each layer in the dense block receives feature maps from all previous layers and passes its output to all subsequent layers. The feature maps received from other layers are merged through cascade. Since the network uses a special connection method of dense connections, the number of layers is reduced. Besides, the reuse of feature maps reduces the network parameters which in turn alleviates the vanishing gradient problem. This idea of efficiently using feature maps between layers was widely used by subsequent researchers.
	\subsection{Graph Convolutional Networks}
	Graph convolutional networks fill the gaps in the development of deep learning that CNNs cannot directly operate non-Euclidean spatial data such as social networks~\cite{kipf2016semi}, protein-protein interaction networks~\cite{fout2017protein}, and knowledge graphs~\cite{hamaguchi2017knowledge}. Similar to the convolution kernel operation of CNN, GCN is also updated by the fusion of information between node neighborhoods. Through the application of convolution operation to the Laplacian matrix of a graph, a connection is established between the spatial domain and the spectral domain~\cite{bruna2013spectral}. This convolution operation is actually a special form of Laplacian smoothing. Therefore, too many layers will cause over smoothing , hence, a GCN network usually has a depth of 3 or 4 layers. As the complexity of the task increases, a basic GCN network cannot meet the requirement. Li et al.~\cite{li2019deepgcns} introduced a deep GCNs by employing the concepts of residual, dense connections and dilated convolutions of CNNs. They designed a 56-layer GCN network that achieved state-of-the-art results. Guo et al.~\cite{li2019deepgcns} designed a graph-to-sequence learning Densely Connected Graph Convolutional Networks (DC-GCNs). This form of GCN combines the local and non-local features of nodes to achieve better graph structure representation in natural language processing (NLP).
	
	\subsection{Graph Convolutional Networks application in 3D mesh}
	\par GCN has also made great progress in the field of 3D mesh. Choi et al.~\cite{choi2020pose2mesh} proposed a GCN network that can extract features from 2D human pose images and generate a 3D human pose mesh in a coarse-to-fine manner. Xin et al.~\cite{wei2020view} proposed a view-based GCN network which uses CNN to extract features from the multi-view picture of the 3D mesh and then constructs a graph structure to accomplish the classification and retrieval tasks by combining CNN and GCN. The above-mentioned work use CNN directly or indirectly to extract the features of the image and generate correlation with GCN. Researchers have developed solutions that exploit the inherent geometric characteristics of 3D surfaces to design different GCNs to accomplish different tasks~\cite{kostrikov2018surface,dominguez2017towards,wu2019active,verma2018feastnet,litany2018deformable}. Ilya Kostrikov et al.~\cite{kostrikov2018surface} designed a GCN network called surface networks by leveraging extrinsic differential geometry properties of 3D surfaces to enhance modelling power. In particular, they used  the Dirac operator, whose spectrum detects principal curvature directions instead of the classical Laplace operator, which directly measures mean curvature. Miguel et al.~\cite{dominguez2017towards} designed a Graph CNN based on graph signal processing theory for 3D point cloud and 3D mesh classification. Wu et al.~\cite{wu2019active} used the 3D mesh face as a unit, and used the geometric features of the face to construct a graph structure through node association. They designed a GCN network through the layer-wise propagation rule to achieve 3D shape co-segmentation, and verify the effectiveness of the algorithm through the COSEG dataset benchmark. Nitika et al.~\cite{verma2018feastnet} designed a novel graph convolution operator to establish the connection between filter weights and arbitrary connected graph neighborhoods which achieved shape representations without relying on shape descriptors. Litany et al.~\cite{verma2018feastnet} designed a variational autoencoder network based on graph convolution operators to complete the deformable shape completion task. The innovation is that it can handle local information of any 3D shape without providing training data for local information and can be applied to any 3D shaped data. Milano et al.~\cite{Milano20NeurIPS-PDMeshNet} designed a primal-dual framework GCN named PD-MeshNet. PD-MeshNet takes the edges and faces of the 3D meshes as the features of the graph nodes, and then adds an attention mechanism to achieve classification and segmentation.
	\section{Method overview}
	The standard GCN network only has 2-3 layers, which is difficult to meet the needs of complex tasks such as 3D object segmentation and classification. Therefore, we need to improve the design of CGN networks. Firstly, we improve the descriptive ability of the graph data by improving the local features of each node, which is achieved through utilizing 1-ring neighborhood spatial and structural features of the mesh face. Secondly, we improve the aggregation ability of the local and non-local features of the GCN network by designing the Densely Connected (DC) GCN block and use it as the key component for constructing effective and efficient GCN application models. 
	\begin{figure}[!]
		\centering
		\resizebox{1.0\textwidth}{!}{
			\includegraphics[height=1.0\linewidth]{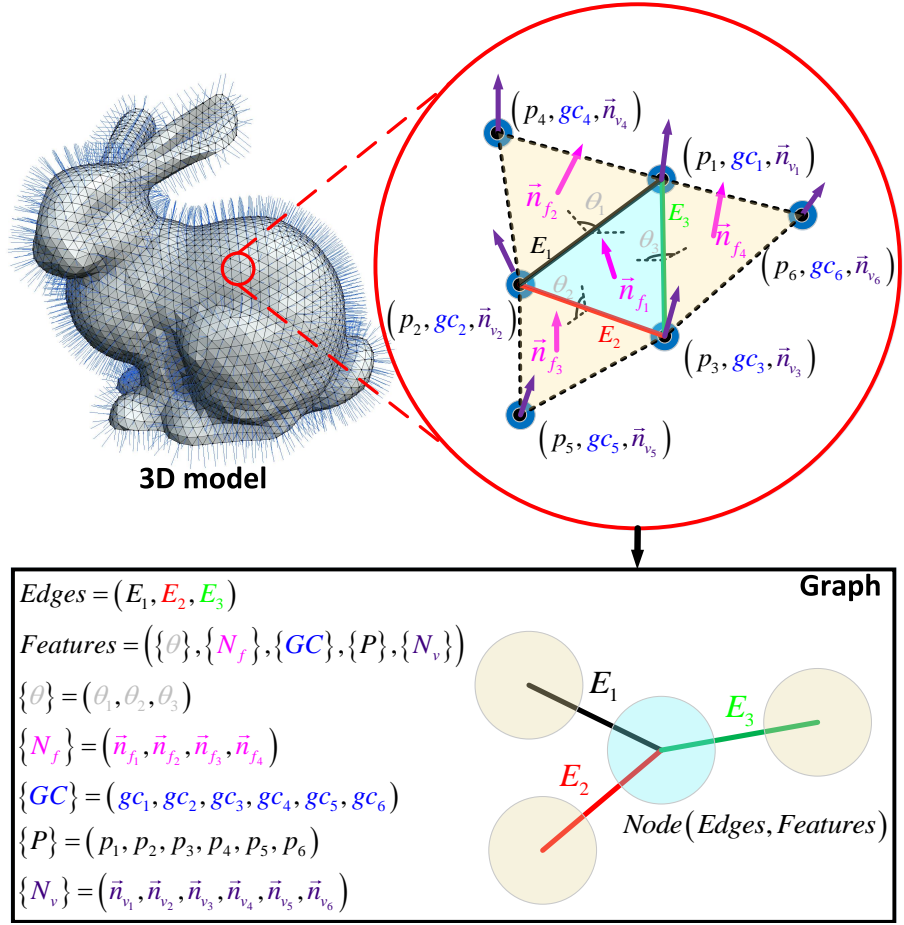}}	
		\caption{ \label{fig:GCN_node} The 1-ring neighbourhood structure of a face on a 3D mesh (top) and the relation between a 3D mesh and its corresponding graph (bottom). Each face in the mesh has a corresponding node in the graph, each edge on the graph represents a connection between a face and one of its 1-ring neighbours.}
	\end{figure}

	\begin{figure}[!]
		\centering
		\resizebox{1.0\textwidth}{!}{
			\subfigure[1-ring neighborhood of a vertex]{\includegraphics[height=0.4\linewidth]{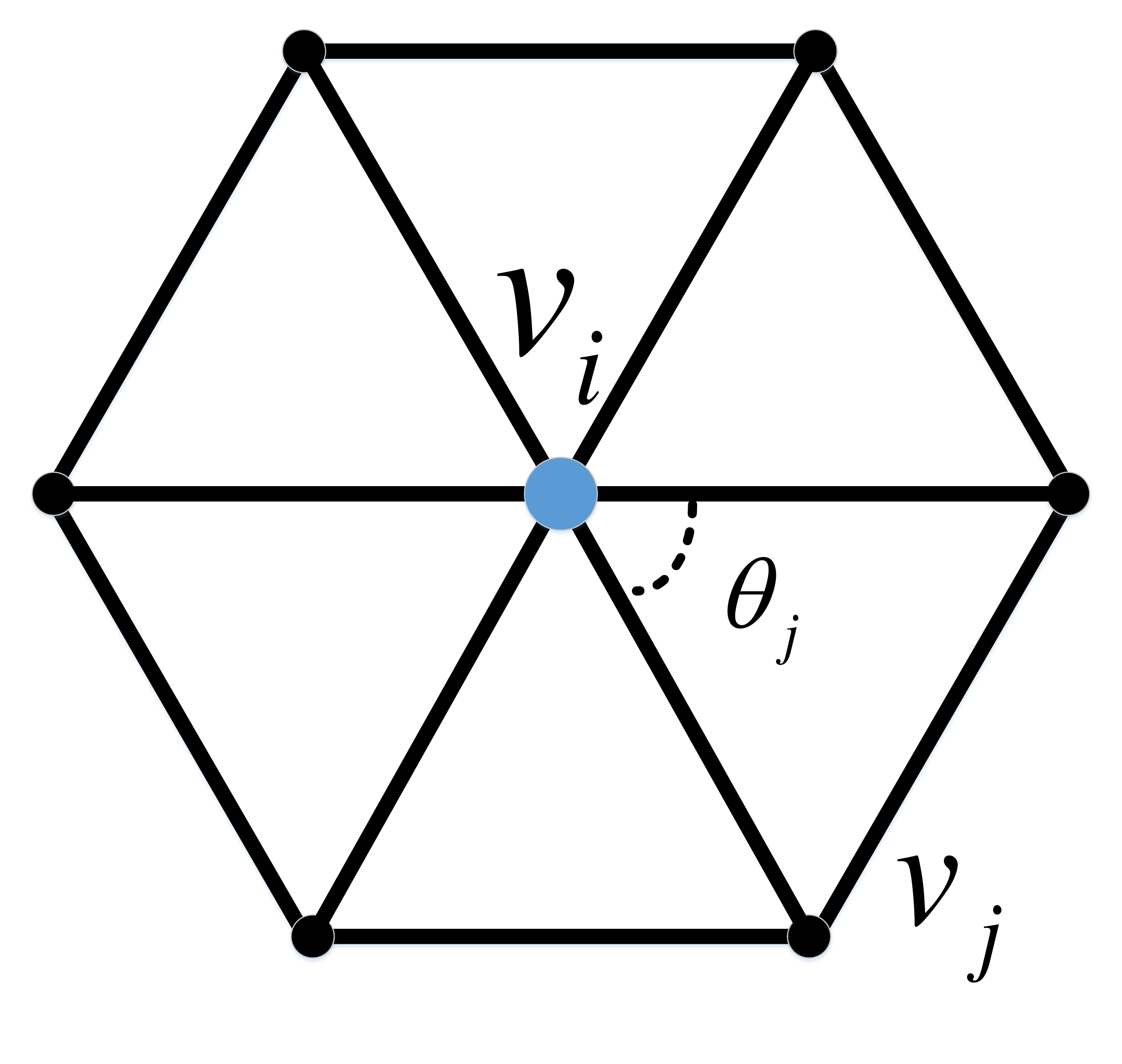}}	
			\subfigure[1-ring neighborhood of a face]{\includegraphics[height=0.4\linewidth]{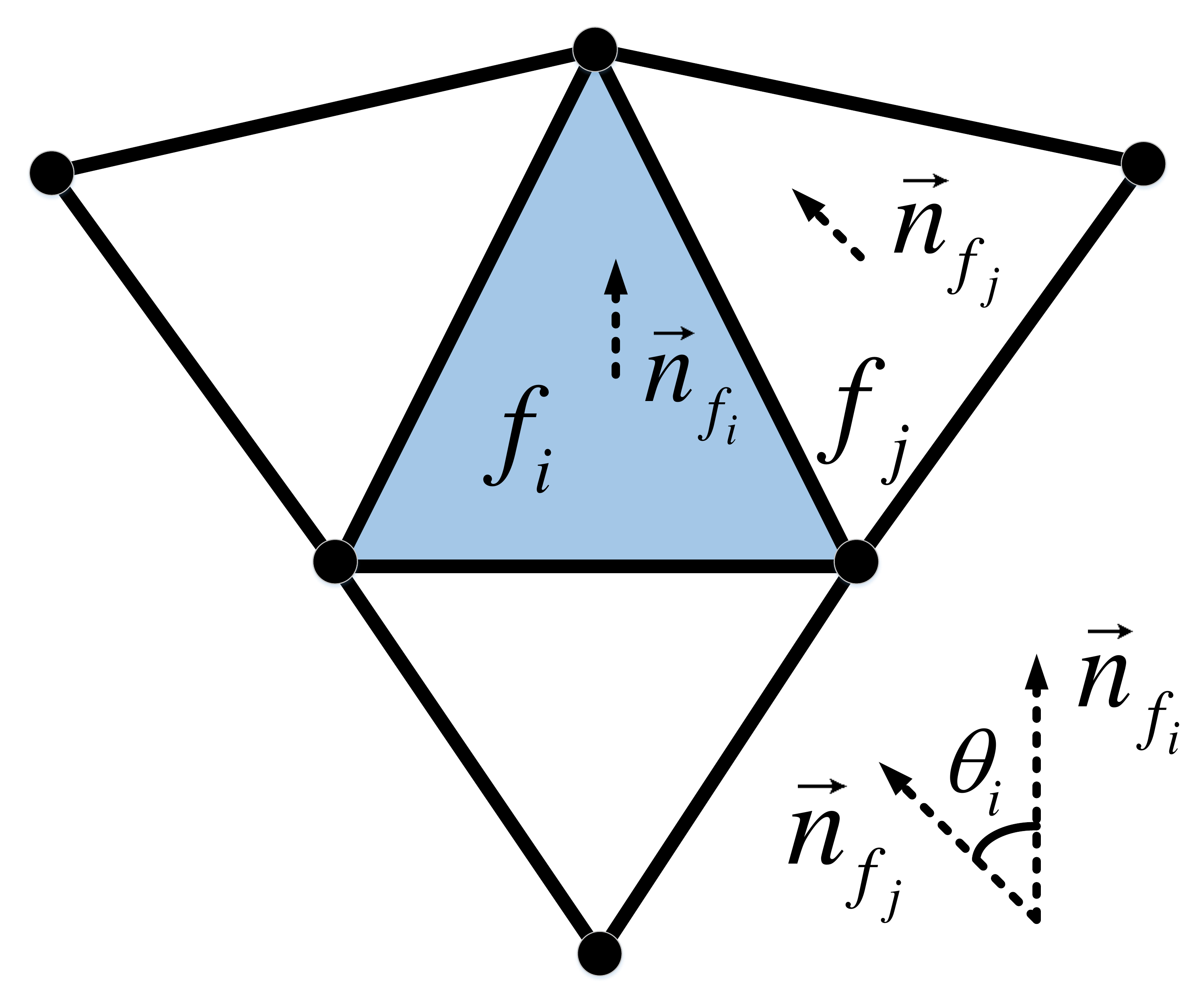}}}	
		\caption{ \label{fig:gaussian_node} 3D triangle meshes.}
	\end{figure}
	\subsection{1-ring neighborhood and node feature vector}
	\par It's well known that in a 3D triangle mesh, each face contains at most three adjacent faces, and the 3D triangle mesh is a non-Euclidean space. The 3D mesh structure data $\mathcal{M}=\{V,~\mathcal{E},~F\}$ of vertices, edges and faces has a clear topology. Therefore, we can convert the 3D triangle mesh into graph data $\mathcal{M}=\{V,~\mathcal{E},~F\}~\Rightarrow~\mathcal{G}=\{N,~E\}$. A face $\{F\}$ on the mesh $\mathcal{M}$ has a corresponding node $\{N\}$ on the graph $\{\mathcal{G}\}$. An edge $\{E\}$ on the graph represents the corresponding edge $\{\mathcal{E}\}$ between a face and one of its 1-ring neighbours on the mesh $\mathcal{M}$. In order to enhance the perception of the local area of the graph node, we use the spatial and structural geometric features ${F}^{(\mathcal{M})}_{faces}$ of the 1-ring neighborhood of the mesh face as the feature $ {F}^{(\mathcal{G})}_{nodes}$ of the graph node, as shown in Eq.~\ref{eq:mesh to graph}:
	\begin{equation}
	\label{eq:mesh to graph}
	{F}^{(\mathcal{G})}_{nodes}={F}^{(\mathcal{M})}_{faces}=\{\{P\},\{N_v\},\{GC\},\{N_f\},\{\theta\}\}
	\end{equation}
	where 1-ring neighborhood spatial and structural features ${F}^{(\mathcal{M})}_{faces}$ (57-dimensional vectors)  include spatial features: the vertex spatial position ($\{P\}$), and structural features: the vertex normal ($\{N_v\}$), Gaussian curvature ($\{GC\}$), the face normal ($\{N_f\}$), angles between the face and its 1-ring neighbourhood faces ($\{\theta\}$), as shown in Fig.~\ref{fig:GCN_node}. The definition of Gaussian curvature on a triangle mesh is via a vertex’s angular deficit~\cite{meyer2003discrete}:
	\begin{equation}
	\label{eq:Gaussian curvature}
	gc(v_i') = (2\pi - \sum_{j\in N(i)} \theta_{j})/A_{ N(i)}
	\end{equation} where $N(i)$ are the triangles incident on vertex $i$ and $\theta_{j}$ is the angle at vertex $i$ in triangle $j$, $A_{ N(i)}$ is the sum of areas of $N(i)$, as shown in Fig.~\ref{fig:gaussian_node} (a). Gaussian curvature is the most important intrinsic geometric quantity in surface theory, and it reflects the degree of curvature of a curved surface. So we introduce it to describe the local features of the mesh. The angle $\theta_{i}$ in the angle set $\{\theta\}$ is defined as the angle between the normals of the two faces (~Eq.~\ref{eq:angle_face_normal}), as shown in Fig.~\ref{fig:gaussian_node} (b). 
	\begin{equation}
	\label{eq:angle_face_normal}
	\theta_{i} = \arccos(\frac{\vec{n}_{f_i}\cdot\vec{n}_{f_j}}{\|\vec{n}_{f_i}\|\cdot\|\vec{n}_{f_j}\|})
	\end{equation}
	The normal $\vec{n}_{v_i}$ in the normal set $\{N_v\}$ is as Eq.~\ref{eq:nvi}~\cite{zhao2006accurate}:
	\begin{equation}\label{eq:nvi}
	\vec{n}_{v_i}=\sum_{j\in F_v(i)} A_j\vec{n}_{f_j},
	\end{equation}
	where $A_j$ is the corresponding face area and $\vec{n}_{f_j}$ is the value of the $j$ face normal, $F_v(i)$ is the number of faces in the $i$-th vertex-ring. The introduction of angles $\{\theta\}$, vertex normals $\{N_v\}$ and face normals $\{N_f\}$ is to enhance the local structural features of the graph nodes, and the vertex space coordinates are used to describe the spatial features of the graph model.
	\par Using 1-ring neighborhood of the faces instead of the vertices as graph nodes has the following advantages:  A triangular face contains three graph, and a face contains more geometric features than a vertex. The 1-ring neighborhood of a triangular face covers a larger area than a vertex, thus enabling nodes to enhance their perception of local areas.


	\subsection{Densely Connected Convolutional Block}
	Graph convolution can directly operate on non-Euclidean space data, similar to image convolution operations on image data. The GCNs convolution operation is obtained by using the weighted average of its neighboring nodes. Kipf et al~\cite{kipf2016semi} proposed a GCN where the node feature convolution operation in the graph $G=\{N,\varepsilon\}$ is defined as:
	\begin{equation}\label{eq:gcn convolutional}
	H_{i}^{(l+1)}=\sigma(\sum_{j\in{\tilde{N}(i)}}H_j^{(l)}W^{(l)}+b^{(l)})
	\end{equation} 
	where $H_{i}^{(l+1)}$ represents the feature of the $i$-th node in the $(l+1)$-th layer, $\sigma$ is a non-linear activation function, $W^{(l)}$ is the weight matrix, $b^{(l)}$ is the bias vector, $\tilde{N}(i)$ is the neighborhood set of node i $\tilde{N}(i)=\{j\in N | (i,j)\in\varepsilon\}$.
	\par Similar to the traditional convolutional neural network of computer vision, as the number of network layers increases, the gradient disappears. The increase in the number of GCN layers will make the effect worse~\cite{kipf2016semi}. If there is no mechanism to restrict the convolutional form of the GCN, then, the best result is 2 layers~\cite{li2018deeper}. However, shallow networks cannot effectively capture non-local features, and have poor applicability to larger graph structures. Bastingset et al.~\cite{bastings2017graph} introduced the resnet mechanism~\cite{he2016deep} into GCNs, as shown in Eq.~\ref{eq:gcn convolutional residual}, each node feature is updated according to Eq.~\ref{eq:gcn convolutional}, and then combined with the updated features of the previous layer.
	\begin{equation}\label{eq:gcn convolutional residual}
	H_{i}^{(l+1)}=\sigma(\sum_{j\in{\tilde{N}(i)}}H_j^{(l)}W^{(l)}+b^{(l)})+H_i^{(l)}
	\end{equation}
	\par Subsequently, researchers have introduced the cyclic layers into the convolutional layer to deepen the network (up to 6 layers)~\cite{guo2019densely}. In order to design a deeper GCN, Guo et al~\cite{guo2019densely} introduced the densenet~\cite{huang2019convolutional} propagation mechanism into GCNs. As shown in Eq.~\ref{eq:gcn h aggregation}, the feature of node $i$ in the $(l+1)$-th layer is not only $H^{(l)}$, but also the set of node features of all previous layers \{$H^{(l)}$,~$H^{(l-1)}$,~...,~$H^{(1)}$\}.
	\begin{equation}\label{eq:gcn h aggregation}
	H_{j}^{(l)}=LA(H_{j}^{(l)},~H_{j}^{(l-1)},~...,~H_{j}^{(1)})
	\end{equation}
	\begin{figure}[!]
		\centering
		\resizebox{1.0\textwidth}{!}{
			\includegraphics[width=0.5\paperwidth]{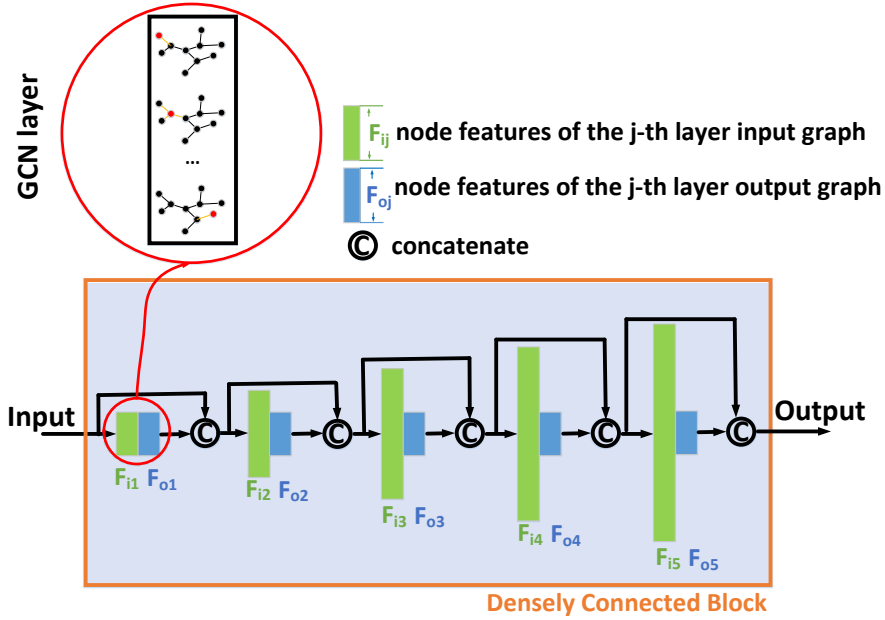} }
		\caption{ \label{fig:dense block} The dense connected block of GCN}
	\end{figure}
	where the LA function can be concatenation, maxpooling or LSTM-attention operations~\cite{xu2018representation}. The advantage of this design is that the number of network layers can be increased deeper, local and non-local features can be better combined, and it is more conducive to improving the description ability of GCNs for graphs.
	\par In 3D mesh classification and segmentation tasks, it is necessary to consider the fusion of local features and non-local features, so we designed a densely connected (DC) GCN block, as shown in Fig.~\ref{fig:dense block}. For DC block specifically, this module is composed of five ordinary GCN layers, and the connection method is realized as Eq.~\ref{eq:gcn h concat} and~\ref{eq:gcn convolutional}.
	\begin{equation}\label{eq:gcn h concat}
	H_{j}^{(l)}=[H_{j}^{(l-1)};~H_{j}^{(l-2)};~...;~H_{j}^{(1)}]
	\end{equation}
	where $[~]$ is a concatenation of the node feature produced in layers $1,~2,~...,~(l-1)$. In order to improve the parameter efficiency similar to DenseNets~\cite{huang2019convolutional}, and we define the feature dimension of each layer of the output graph node in the DC block to be equal, as show in Eq.~\ref{eq:gcn deatures},
	\begin{equation}\label{eq:gcn deatures}
	DIM(H_{i(out)}^{(1)}) = DIM(H_{i(out)}^{(2)})= . . . = DIM(H_{i(out)}^{(n)}) = \tau
	\end{equation}
	where $H_{i(out)}^{(1)}$ represents the output feature of the $i$-th node of the first layer, $DIM(X)$ represents the dimensionality of the vector $X$, and $\tau$ can be freely set according to different tasks by the users. Therefor the node feature growth rate could be  defined as follow: 
	\begin{equation}\label{eq:gcn node features}
	DIM(H_{i(input)}^{(l)})=(n-1)*DIM(H_{i(out)}^{(l-1)}) + DIM(H_{i(out)}^{(1)}) .
	\end{equation}
	In order to effectively use the parameters and prevent the feature dimension of GCNs from being too wide, $n$ cannot be set too large. Our experiments empirically set $n~=~5$. As shown in Fig.~\ref{fig:dense block}, the dimension of the input feature of the 5-th layer of the GCN in the DC block is~$DIM(H_{i(input)}^{(5)})=(5-1)*\tau+\tau=5\tau$. It should be noted that the dimension of the DC input is $\tau$ and that of the output is $6\tau$.
	\subsection{MDC-GCN for 3D Mesh Classification}
	\par
\begin{figure}[htbp]
		\centering
		\resizebox{1.0\textwidth}{!}{
			\includegraphics[width=1.0\paperwidth]{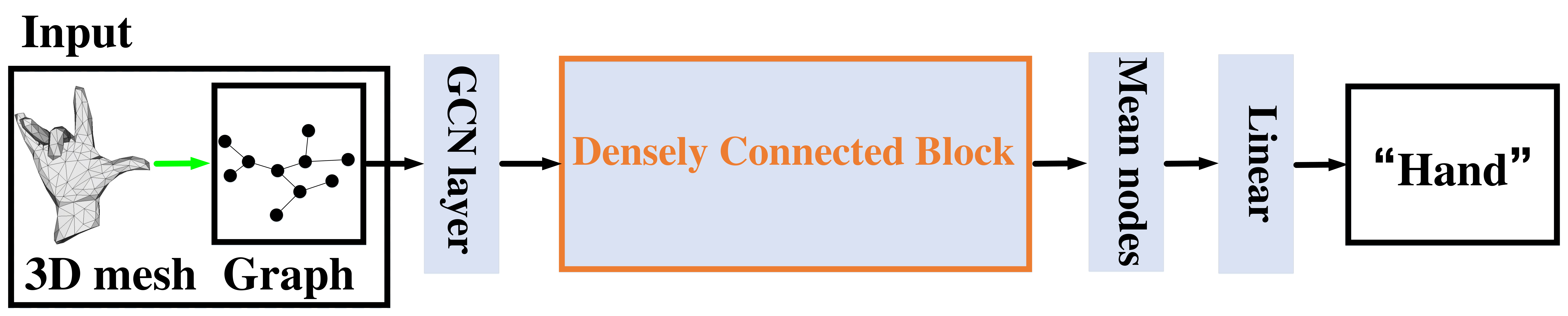} }
		\caption{ \label{fig:MDC_GCN_network_claasification} MDC-GCN for 3D mesh classification.}
	\end{figure}

 The proposed MDC-GCN for 3D mesh classification is composed of a DC block, a GCN layer, a mean graph nodes layer, and a linear layer as illustrated in Fig.~\ref{fig:MDC_GCN_network_claasification}~. An ordinary GCN layer is used as the terminal layer between the DC block and the input. Through this layer, we can get the features of the input graph node defined by the DC block. After the feature extraction operation of the graph convolution of a DC block, the feature average value of the graph node is obtained through a layer of graph mean nodes, and finally the classification category is output through the linear layer. The parameter design of the MDC-GCN for 3D mesh classification is shown in Table~\ref{fig:classification model}.
	\par With the help of the multi-dimensional features of 1-ring structure of a single node and the design of the DC block that aggregates local and non-local features, MDC-GCN only uses 8 GCN layers to complete the 3D mesh classification task. The experiments and ablation experiments in the following section verify that the modular design of MDC-GCN allows users to easily change the parameters, which is very simple and efficient.
	
	\begin{table}[h] 
		\centering  
		\caption{MDC-GCN classification network configuration.} 
		\label{fig:classification model}
		\begin{tabular}{c|c cl}  
			\cmidrule{1-2}
			Parameters&GCN layer\\
			\cmidrule{1-2}
			$(in=\textbf{input},~out=1024)$&1\\
			$(in=1024,~out=1024)$&2& \hspace{-2.5em}\rdelim\}{5}{*}[DC block] \\
			$(in=2048,~out=1024)$&3\\
			$(in=3072,~out=1024)$&4\\
			$(in=4096,~out=1024)$&5\\
			$(in=5120,~out=1024)$&6\\
			$(graph ~mean~ nodes)$&7\\
			$(in=6144,~out=\textbf{output})$&8\\
			\cmidrule{1-2}
		\end{tabular}
	\end{table}
	
	\subsection{MDC-GCN for 3D Mesh Segmentation}
	\par In the deep learning literature of 2D computer vision, the segmentation network is often much more complicated than the classification network. The 3D mesh segmentation task can be regarded as a multi-classification task, and its complexity is far greater than that of 3D mesh classification, so more parameters are needed to meet the needs of the segmentation task. Different from the design of many deep GCN networks, the efficient local feature description of the graph data and the design of the DC block enable MDC-GCN to complete the 3D mesh segmentation with very few GCN layers. The design of 2 DC blocks instead of more or fewer blocks is to achieve balance between performance and parameters size. See Table~\ref{fig:Human Body Segmentation_parameters}.
	\par
\begin{figure}[htbp]
		\centering
		\resizebox{1.0\textwidth}{!}{
			\includegraphics[width=1.0\paperwidth]{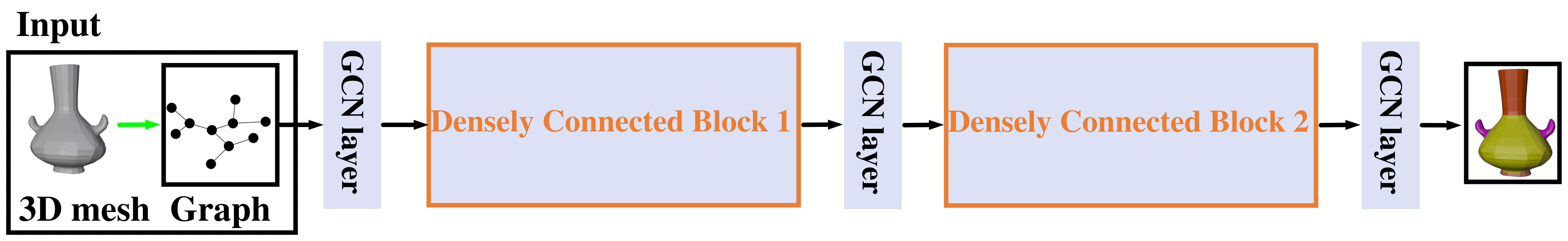} }
		\caption{ \label{fig:MDC_GCN_network_segmentation} MDC-GCN for 3D mesh segmentation.}
	\end{figure}

 The proposed MDC-GCN for 3D mesh segmentation is composed of 2 DC blocks and 3 GCN layers as illustrated in Fig.~\ref{fig:MDC_GCN_network_segmentation}~. Different from the classification task, we introduce two DC blocks and use a GCN as the intermediate connection layer for feature extraction. Finally, a layer of GCN is used as the output layer, and the dimension of its output is the segmentation category. The parameter design of the MDC-GCN for 3D mesh segmentation is shown in Table~\ref{fig:segmentation model}.
	\begin{table}[h] 
		\centering 
		\caption{MDC-GCN segmentation network configuration.} 
		\label{fig:segmentation model}
		\begin{tabular}{c|c cl} 
			\cmidrule{1-2}
			Parameters&GCN layer\\
			\cmidrule{1-2}
			$(in=\textbf{input},~out=1024)$&1\\
			$(in=1024,~out=1024)$&2& \hspace{-2.5em}\rdelim\}{5}{*}[DC block 1] \\
			$(in=2048,~out=1024)$&3\\
			$(in=3072,~out=1024)$&4\\
			$(in=4096,~out=1024)$&5\\
			$(in=5120,~out=1024)$&6\\
			$(in=6144,~out=1024)$&7\\
			$(in=1024,~out=1024)$&8& \hspace{-2.5em}\rdelim\}{5}{*}[DC block 2] \\
			$(in=2048,~out=1024)$&9\\
			$(in=3072,~out=1024)$&10\\
			$(in=4096,~out=1024)$&11\\
			$(in=5120,~out=1024)$&12\\
			$(in=6144,~out=\textbf{output})$&13\\
			\cmidrule{1-2}
		\end{tabular}
	\end{table}
	

	\subsection{Optimizer, loss function and implementation platform}
	The classification target variables of MDC-GCNs are discrete so we use cross entropy to define loss function. Segmentation can also be seen as a multi-classification task. Hence, the loss function of MDC-GCNs can be described as:
	\begin{equation}\label{eq:loss}
	loss(x, class) = -\log\left(\frac{\exp(x[class])}{\sum_j \exp(x[j])}\right)= -x[class] + \log\left(\sum_j \exp(x[j])\right)
	\end{equation}
	where $x$, $class$, $j$ are predicted vector, ground truth label and the $j$-th value of $x$, respectively. In the selection of MDC-GCN optimizer, we use the Adam optimizer proposed by Kingma et al~\cite{kingma2014adam}.
	\par We use the open source efficient graph neural network framework (Deep Graph Library, DGL) proposed by Want et al ~\cite{kingma2014adam} to implement the MDC-GCNs.  Our experimental platform: windows 10, NVIDIA GTX 1080 graphics card, Intel Xeon quad-core 3.7 Ghz CPU, and 96 G RAM.

	
	\section{Experiments}
	\par We have conducted comparative analysis of multiple sets of experiments of applying MDC-GCNs to the problems of segmenting 3D shapes and classifying 3D objects represented in 3D meshes. We will first explain the datasets used in the experiments and then present experimental results of 3D mesh segmentation and classification respectively.
	\subsection{Datasets and Data Preprocessing}
	We use the open source datasets from~\cite{hanocka2019meshcnn} to test the MDC-GCNs’ performances in 3D mesh segmentation and classification. As mentioned earlier, we use the triangular faces of the 3D mesh as the basic processing units, which is similar to the pixels of the image. In order to train MDC-GCNs more effectively, we resize the meshes in the dataset to a mesh with a fixed number of faces. As in meshCNN, we don't need to fix the number of input mesh faces and can input meshes of any resolution. Unlike meshCNN, our smallest unit is a triangle face instead of the edge of a triangle face. We use triangular faces as the unit, and use the spatial and structural features of its 1-ring neighborhood to convert the mesh into the input graph structure data of MDC-GCNs, as shown in Fig.~\ref{fig:GCN_node}. It is worth noting that MDC-GCNs do not rely on data augmentation. 
	
	\subsection{Classification}
	\par
\begin{figure}[htbp]
		\centering
		\resizebox{1.0\textwidth}{!}{
			\includegraphics[height=0.4\linewidth]{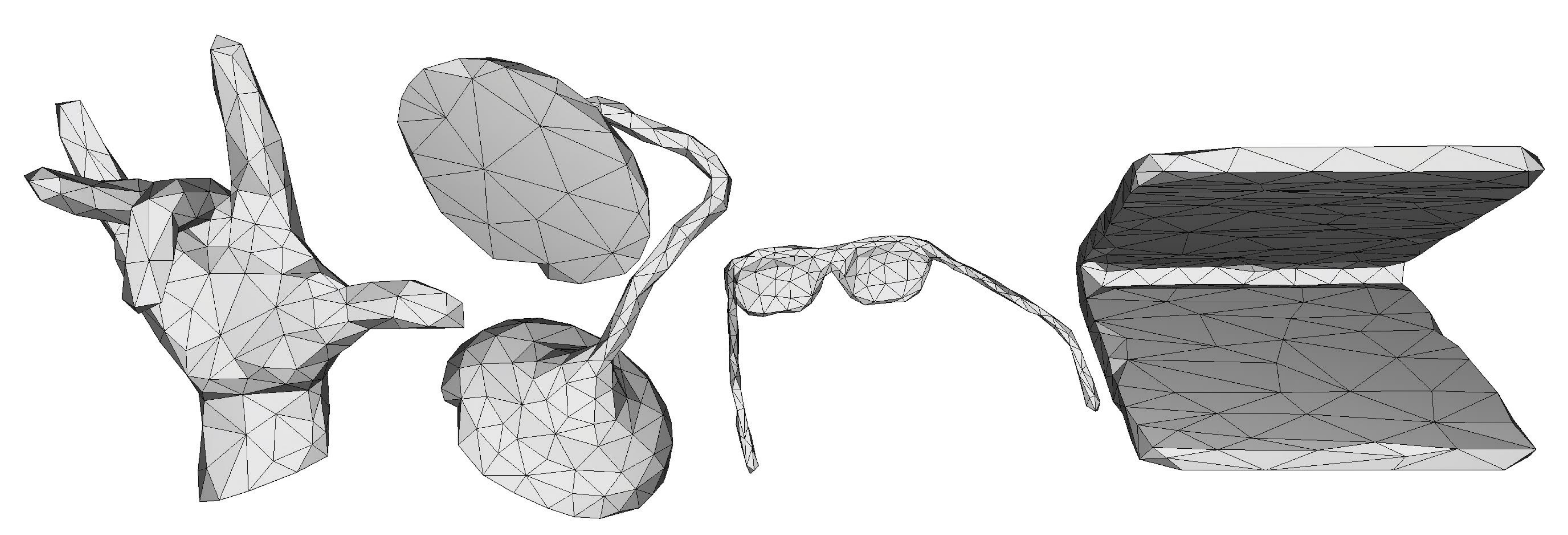}
		}
		\caption{ \label{fig:sherec} The four meshes in SHREC\cite{bronstein2011shape}, from left to right are hand, lamp, glasses, and notebook, respectively.}
	\end{figure}

 In the SHEREC dataset, there are 30 different types of rigid and non-rigid watertight meshes. Fig.~\ref{fig:sherec} shows four types of mesh. Each type contains 20 meshes of the same type but with different shapes. Each mesh contains 500 faces, and each face is used as a graph node. Each node contains $57$-dimensional feature vectors. If the input batchsize is $B$, then the input data dimension is $(B\times500\times57)$. Similar to GWCNN~\cite{ezuz2017gwcnn} and~meshCNN~\cite{hanocka2019meshcnn}, we divide $80\%$ (16) and $50\%$ (10) of each category into two training sets. We also follow the random division mechanism of meshCNN. Our result is the average value obtained from $5$ randomly generated training sets. The learning rate $lr=0.0003$, $droupout=0.3$, and the optimizer is Adam. Comparison with methods in the literature is shown in Table~\ref{fig:SHREC_results}.
	
	\begin{table}[h]  
		\centering  
		\caption{Results on SHREC dataset}  
		\label{fig:SHREC_results}
		\begin{tabular}{ccc}  
			\hline
			Algorithm & Accuracy~(Split 16)& Accuracy~(Split 10) \\ [0.5ex]  
			\hline
			Our & \textbf{99.7$\%$} & \textbf{99.2$\%$} \\
			Milano et al.~\cite{Milano20NeurIPS-PDMeshNet} & 99.7$\%$ & 99.1$\%$ \\
			Hanocka et al.~\cite{hanocka2019meshcnn} & 98.6$\%$ & 91.0$\%$ \\
			Ezuz et al.~\cite{ezuz2017gwcnn} & 96.6$\%$ & 90.3$\%$ \\
			Sinha et al.~\cite{sinha2016deep} & 96.6$\%$ & 88.6$\%$ \\
			Zhirong et al.~\cite{wu20153d} & 48.4$\%$ & 52.7$\%$ \\
			Bronstein et al.~\cite{bronstein2011shape} & 70.8$\%$ & 62.6$\%$ \\
			\hline
		\end{tabular}
	\end{table}

	\par
\begin{figure}[htbp]
		\centering
		\includegraphics[height=0.5\linewidth]{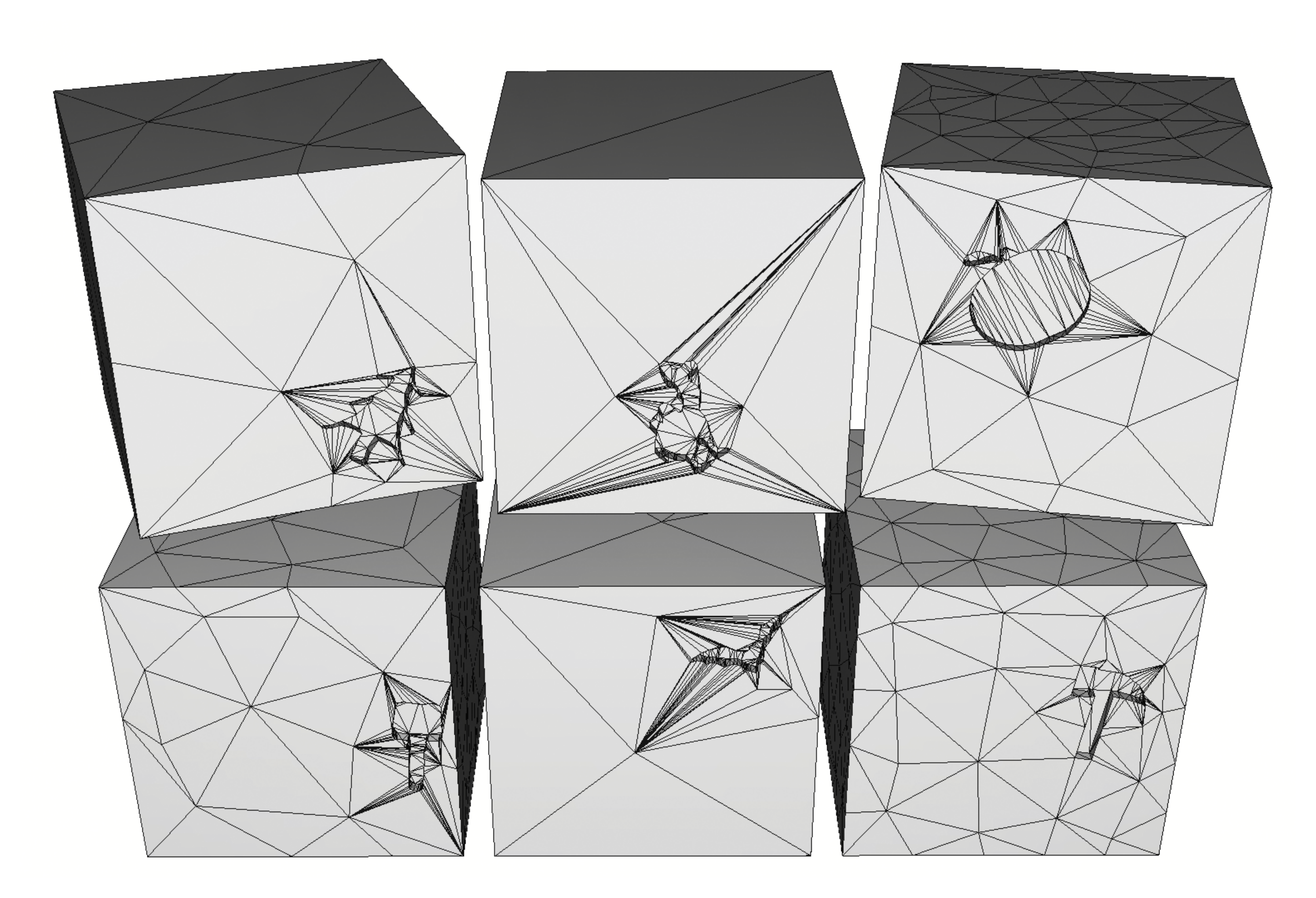}	
		\caption{ \label{fig:cube} The four meshes in Cube Engraving dataset~\cite{latecki2000shape}, from left to right in the top row are tree, camel, and apple, respectively. From left to right in the bottom row are key, bat, hammer, respectively.}
	\end{figure}

 We have also conducted experiments on the Cube Engraving dataset. This dataset comes from ~\cite{latecki2000shape} was also used in meshCNN. Six meshes of the Cube Engraving dataset are shown in Fig.~\ref{fig:cube}. The special feature of this data set lies in a model containing 500 faces, most of which are planes, and only a small part of the small holes are the shapes that we need to recognize. Moreover, the position of the shape in each cube is random, classification on this dataset tests the 3D shape learning ability of MDC-GCNs to the extreme. The parameters and optimizer are the same as the previous classification experiment (SHEREC). Comparison with methods in the literature is shown in Table~\ref{fig:cubes_results}.
	
	\begin{table}[h]  
		\centering  
		\caption{Results on Cube Engraving dataset}  
		\label{fig:cubes_results}
		\begin{tabular}{cc}  
			\hline
			Algorithm & Accuracy \\ [0.5ex]  
			\hline
			Our & \textbf{94.99$\%$} \\
			Milano et al.~\cite{Milano20NeurIPS-PDMeshNet} & 94.39$\%$ \\
			Hanocka et al.~\cite{hanocka2019meshcnn} & 92.16$\%$ \\
			Charles et al.~\cite{qi2017pointnet++} & 64.26$\%$ \\
			\hline
		\end{tabular}
	\end{table}
	
	\begin{figure}[htbp]
		\centering
		\subfigure[Telealiens (Ours)]{
			\includegraphics[width=5.5cm]{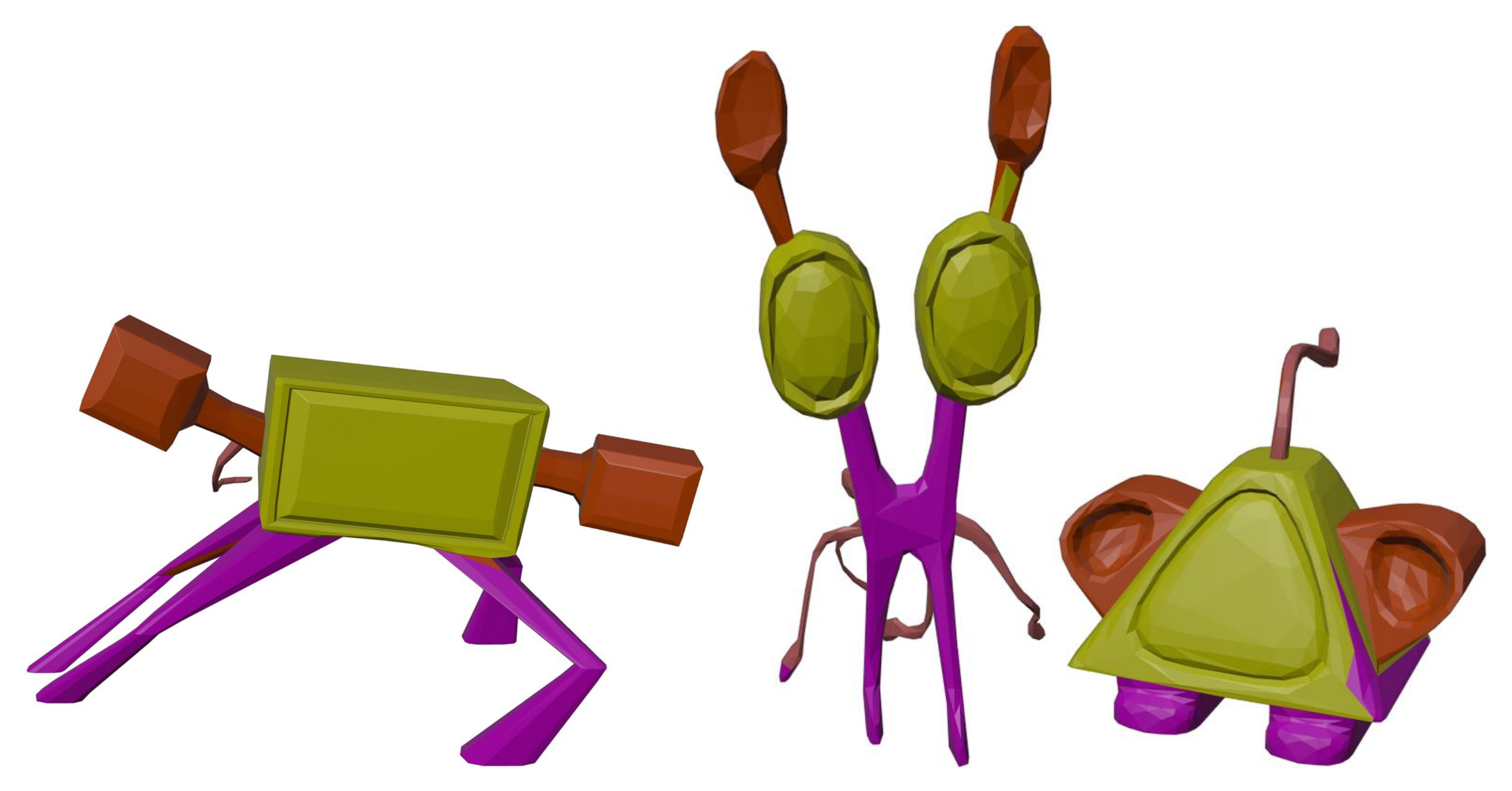}}
		\subfigure[Telealiens (Ground truth)]{
			\includegraphics[width=5.5cm]{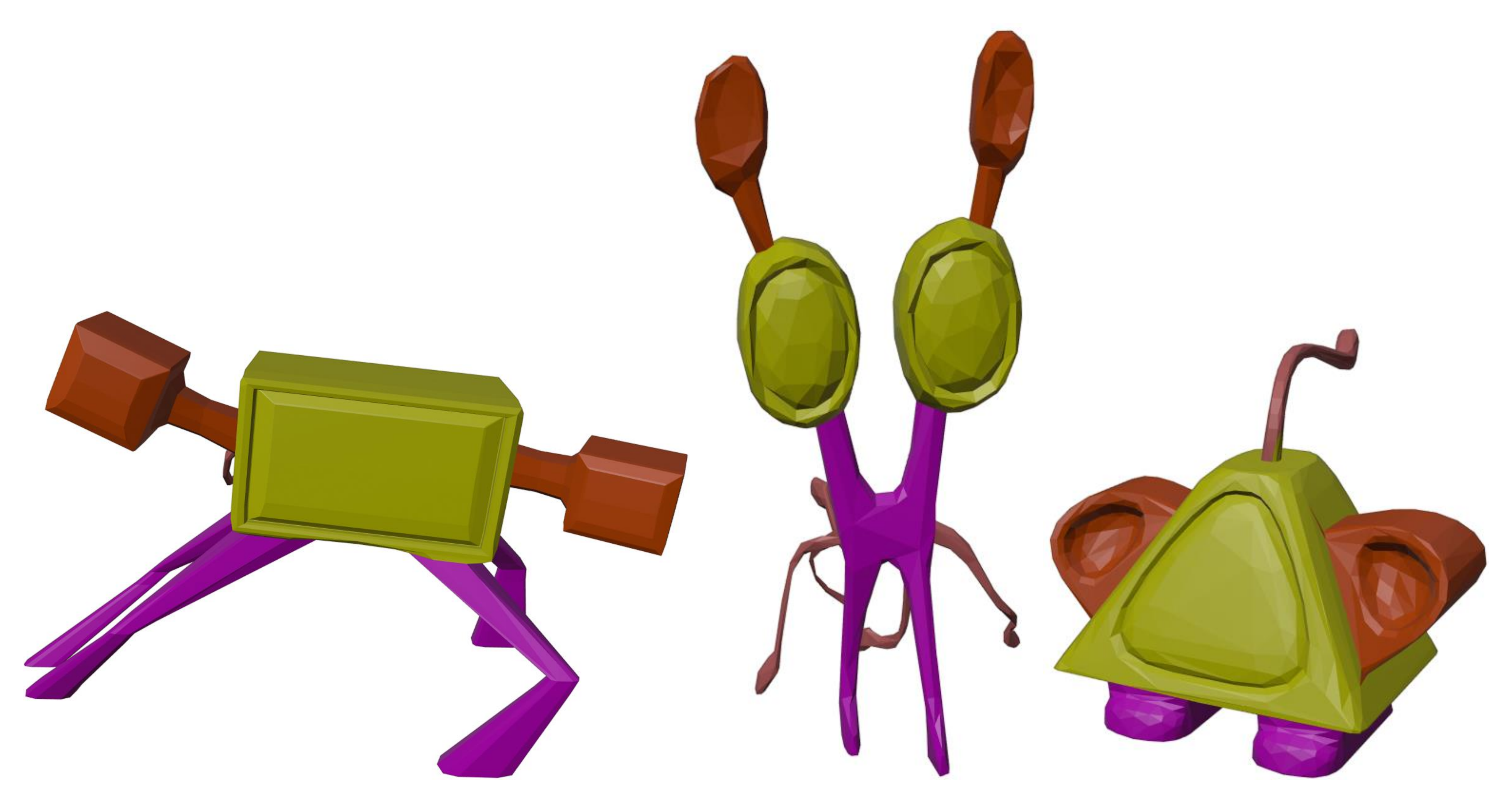}}
		\subfigure[Vases (Ours)]{
			\includegraphics[width=5.5cm]{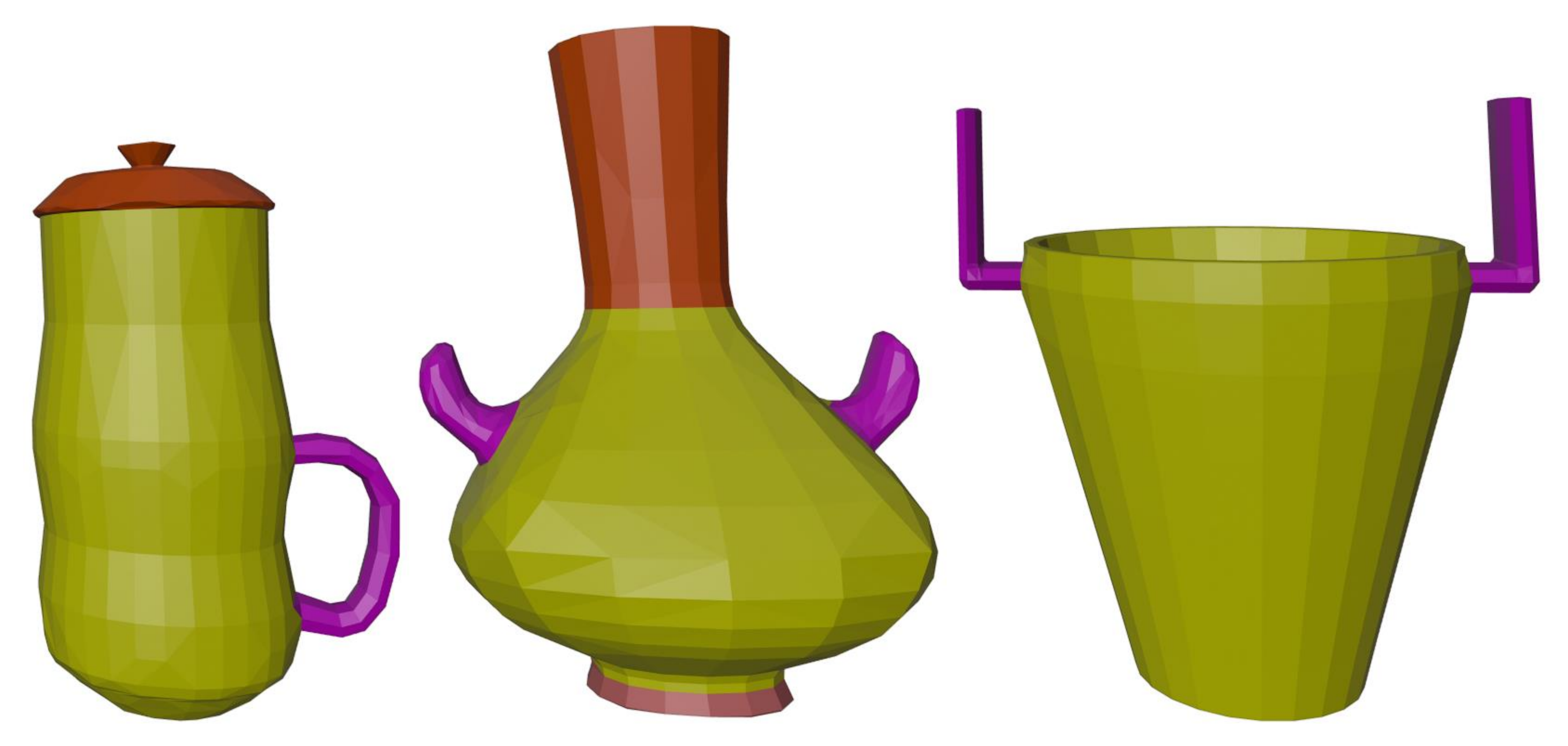}}
		\subfigure[Vases (Ground truth)]{
			\includegraphics[width=5.5cm]{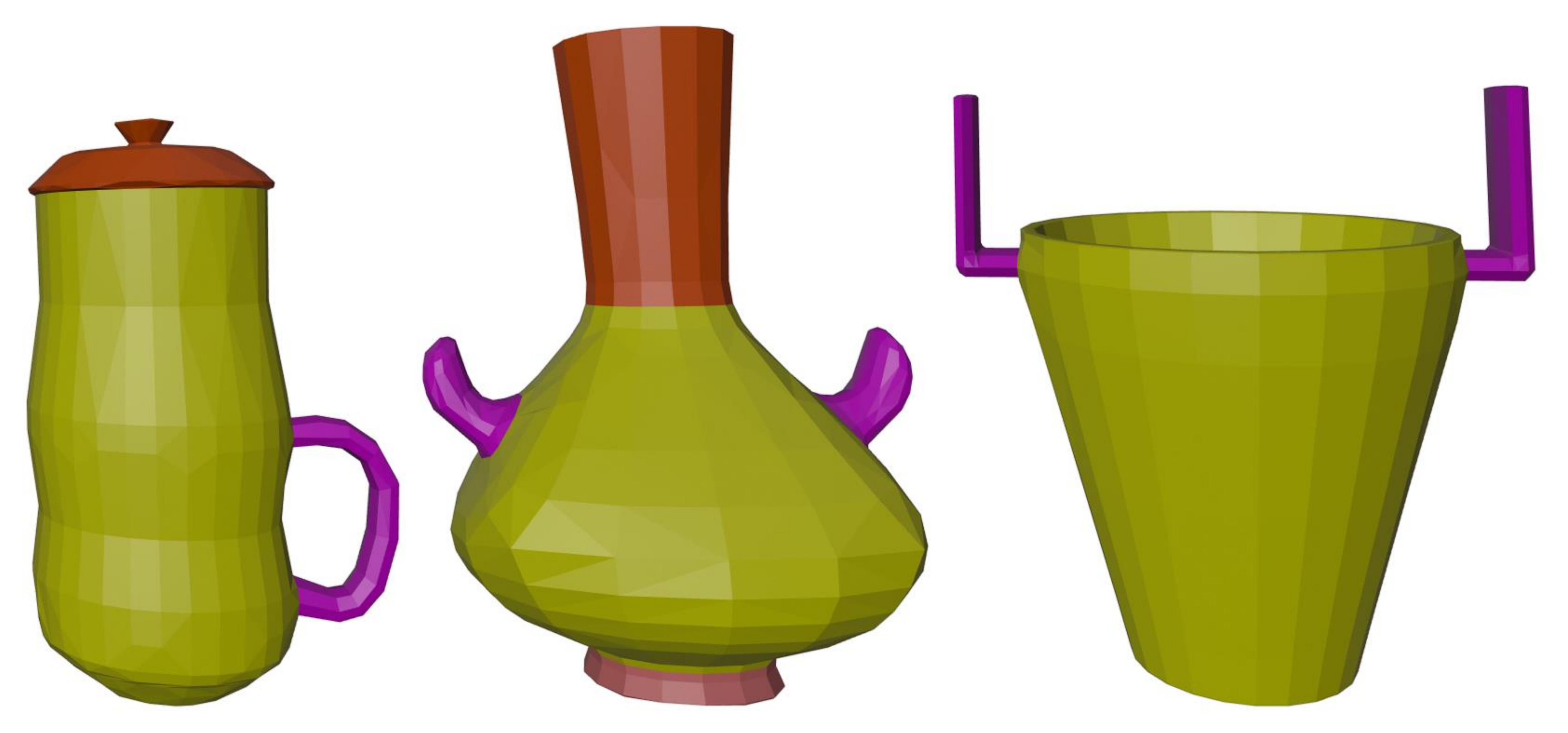}}
		\subfigure[Chairs (Ours)]{
			\includegraphics[width=5.5cm]{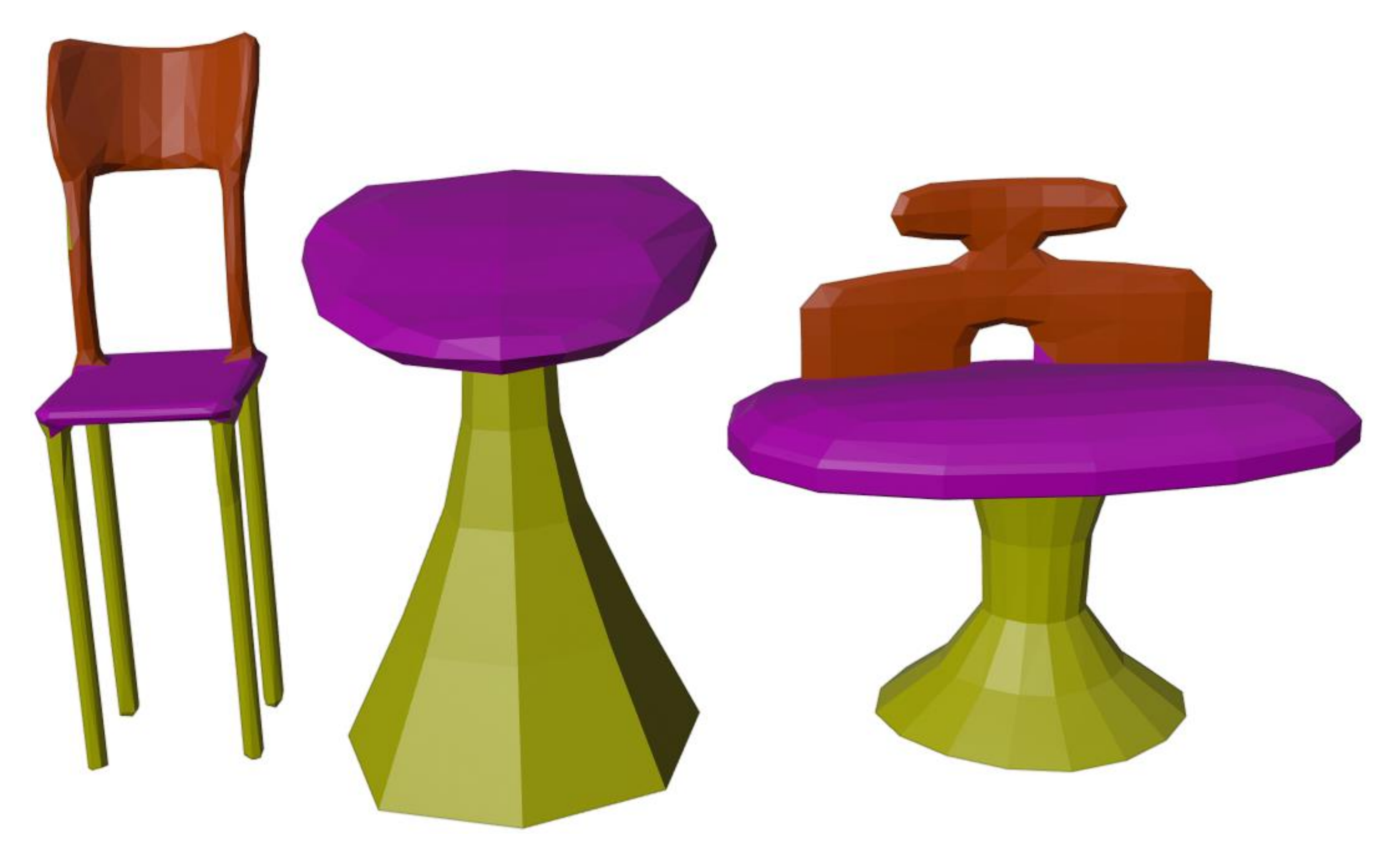}}
		\subfigure[Chairs (Ground truth)]{
			\includegraphics[width=5.5cm]{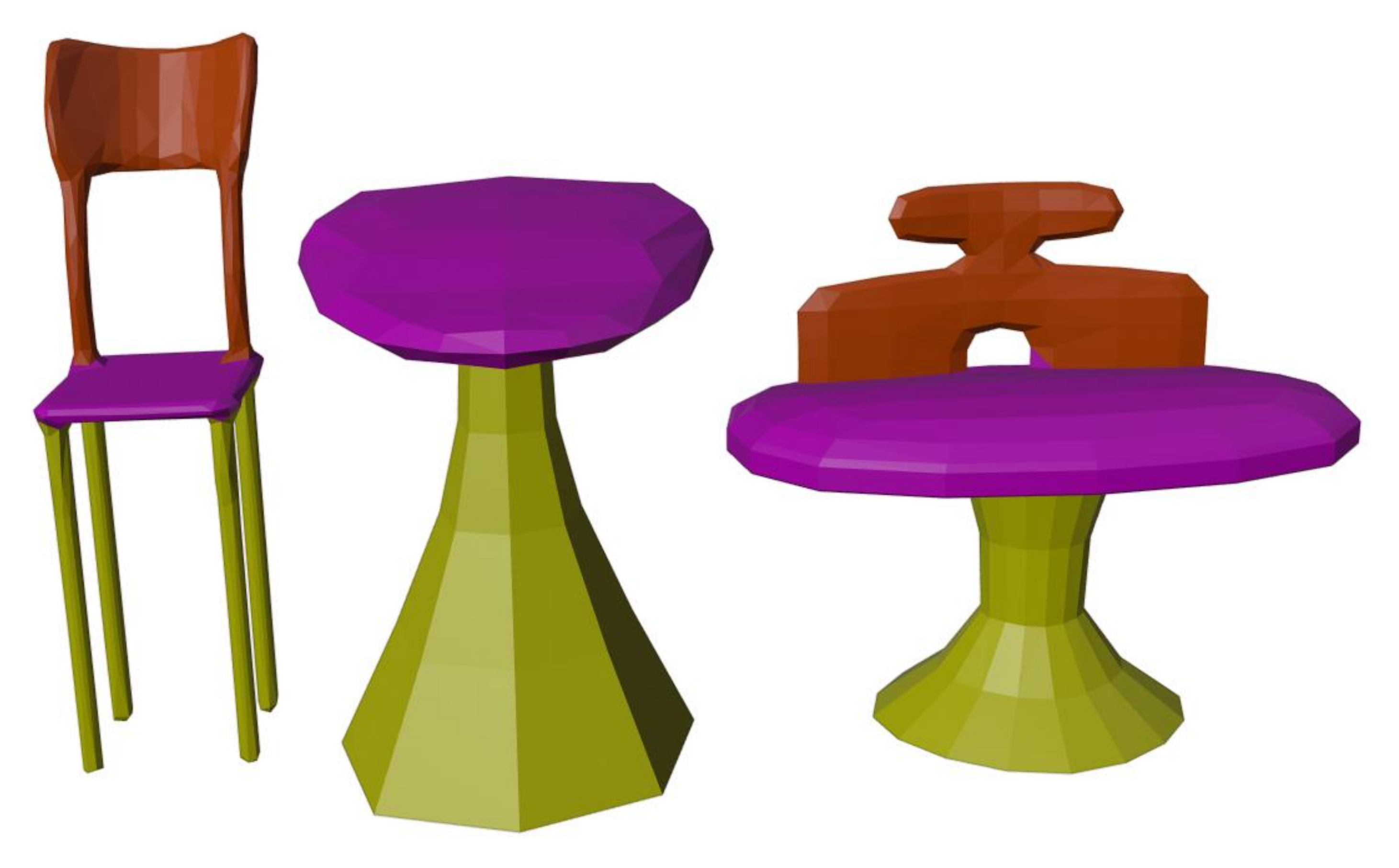}}
		\subfigure[Human Body (Ours)]{
			\includegraphics[width=5.5cm]{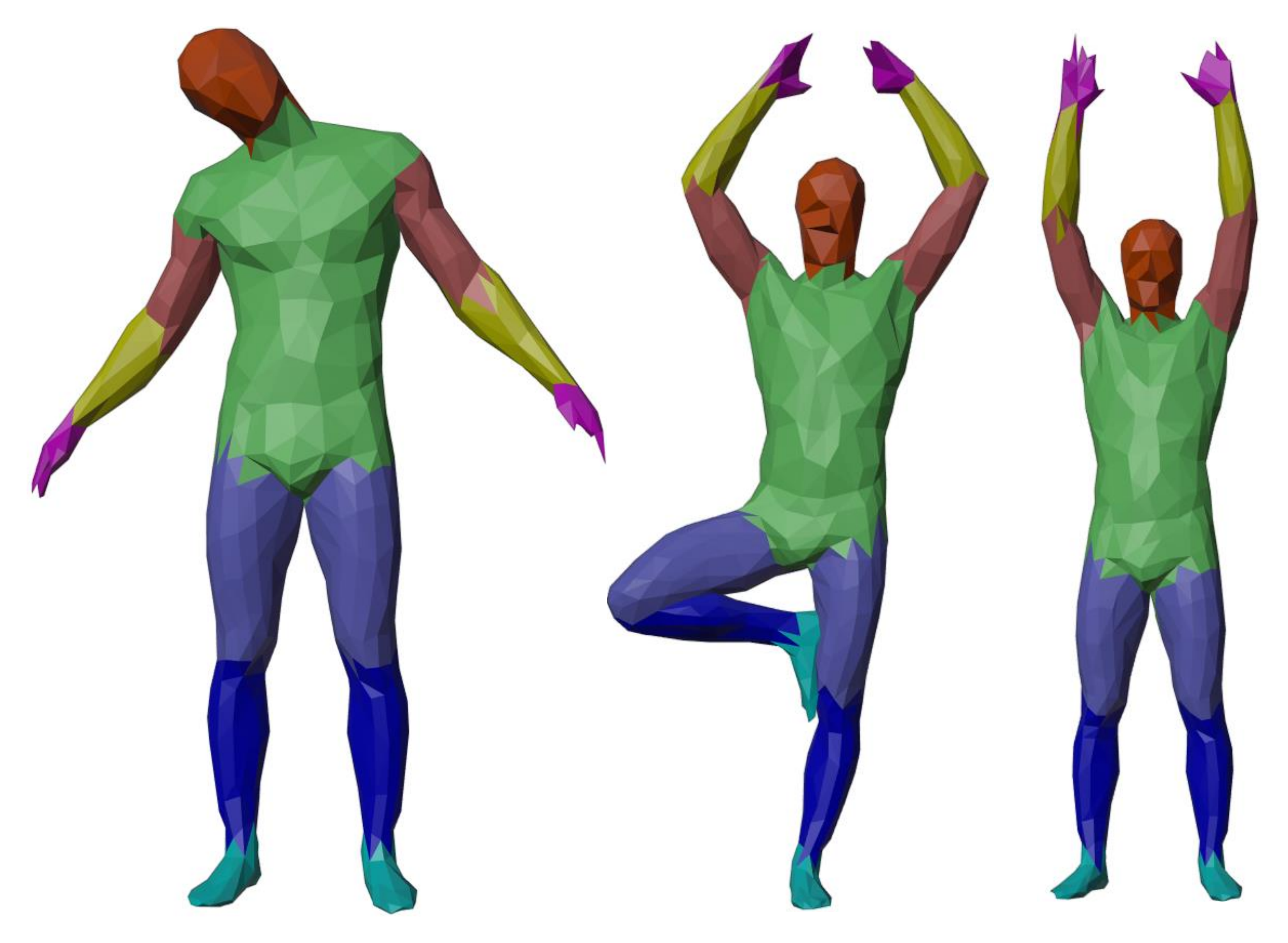}}
		\subfigure[Human Body (Ground truth)]{
			\includegraphics[width=5.5cm]{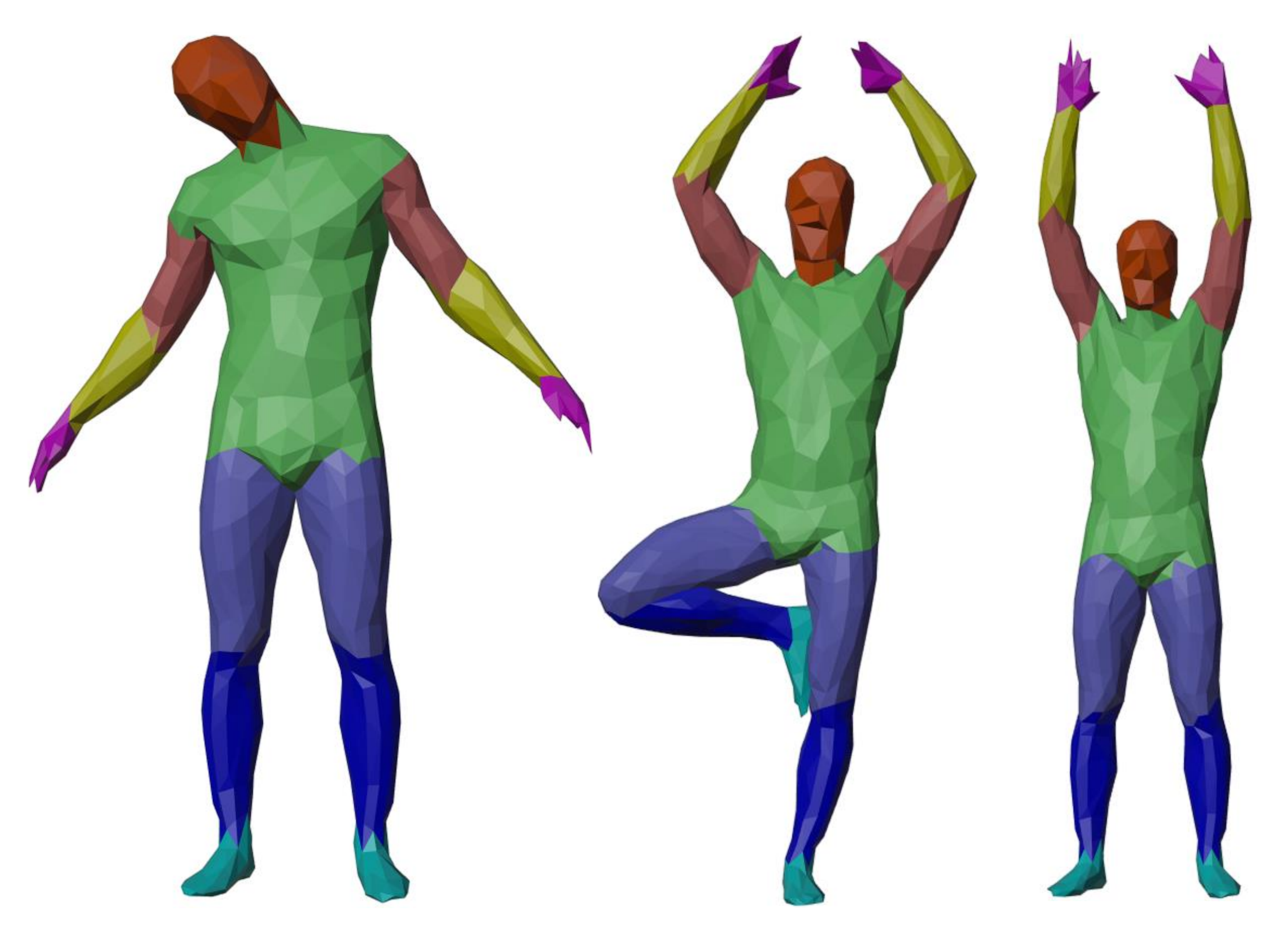}}
		\caption{Visualization of segmentation results with face labels. (a), (c), (e), (f) are the prediction results of MDC-GCNs on the Telealiens, Vases, Chairs, and Human Body data sets respectively (three randomly selected). (b), (d), (f), (g) are the corresponding ground truth. }
		\label{fig:aliens_segmentation}
	\end{figure}
	
	\begin{table}[h]  
		\centering  
		\caption{Results on COSEG segmentation}  
		\label{fig:COSEG_results}
		\resizebox{1.0\textwidth}{!}{
			\begin{tabular}{cccc}  
				\hline
				Algorithm & Vases Acc&  Chairs Acc& Telealiens Acc \\ [0.5ex]  
				\hline
				Our & 97.68$\%$ & \textbf{99.74$\%$} & 95.78$\%$ \\
				Milano et al.~\cite{Milano20NeurIPS-PDMeshNet} & \textbf{97.83$\%$} & 98.21$\%$ & \textbf{99.03$\%$} \\
				Hanocka et al.~\cite{hanocka2019meshcnn} & 97.27$\%$ & 99.63$\%$ & 97.56$\%$ \\
				Charles et al.~\cite{qi2017pointnet} & 91.5$\%$ & 70.2$\%$ & 54.4$\%$ \\
				Charles et al.~\cite{qi2017pointnet++} & 94.7$\%$ & 98.9$\%$ & 79.1$\%$ \\
				Yangyan et al.~\cite{li2018pointcnn} & 96.37$\%$ & 99.31$\%$ & 97.40$\%$ \\
				\hline
			\end{tabular}
		}
	\end{table}

	\begin{table}[h]  
		\centering  
		\caption{Results on Human Body Segmentation}  
		\label{fig:Human Body Segmentation_results}
		\resizebox{1.0\textwidth}{!}{
			\begin{tabular}{ccc}  
				\hline
				Feature & Dimensions& Acc \\ [0.5ex]  
				\hline
				Our & 57 & \textbf{94.65$\%$}  \\
				Milano et al.~\cite{Milano20NeurIPS-PDMeshNet} & 5  & 91.11$\%$ \\
				Hanocka et al.~\cite{hanocka2019meshcnn} & 5  & 92.30$\%$ \\
				Haim et al.~\cite{haim2019surface} & 3  & 91.31$\%$ \\
				Maron et al.~\cite{maron2017convolutional} & 26  & 88.00$\%$ \\
				Charles et al.~\cite{qi2017pointnet++} & 3  & 90.77$\%$ \\
				Yue et al.~\cite{wang2019dynamic} & 3  & 89.72$\%$ \\
				Masci et al.~\cite{masci2015geodesic} & 64  & 86.40$\%$ \\
				Poulenard and Ovsjanikov~\cite{poulenard2019multi} & 64  & 89.47$\%$ \\
				\hline
			\end{tabular}
		}
	\end{table}

	\begin{table}[h]  
		\centering  
		\caption{Comparison of MDC-GCN segmentation network parameters with PD-MeshNet and MeshCNN.}  
		\label{fig:Human Body Segmentation_parameters}
		\resizebox{1.0\textwidth}{!}{
			\begin{tabular}{ccc}  
				\hline
				Method & Parameters& Acc \\ [0.5ex]  
				\hline
				Our~(MDC-GCN) & \textbf{147,828} & \textbf{94.65$\%$}  \\
				Milano et al.~\cite{Milano20NeurIPS-PDMeshNet}~(PD-MeshNet,~2020) & 173,728  & 91.11$\%$ \\
				Hanocka et al.~\cite{hanocka2019meshcnn}~(MeshCNN,~2019) & 2,279,720  & 92.30$\%$ \\
				\hline
			\end{tabular}
		}
	\end{table}

	\begin{figure}[htbp]
		\centering
		\subfigure[Telealiens (Ours)]{
			\includegraphics[width=5.5cm]{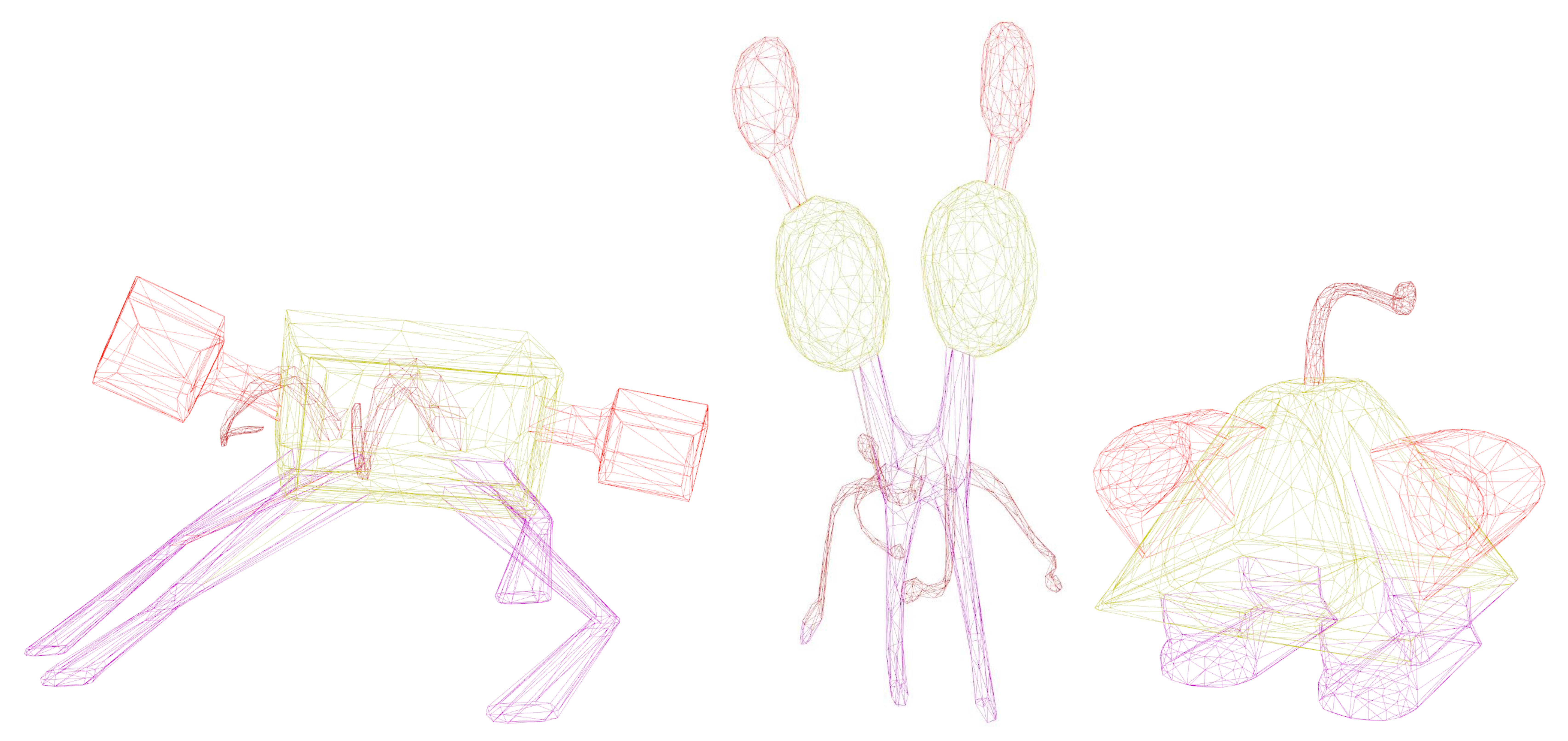}}
		\subfigure[Telealiens (Ground truth)]{
			\includegraphics[width=5.5cm]{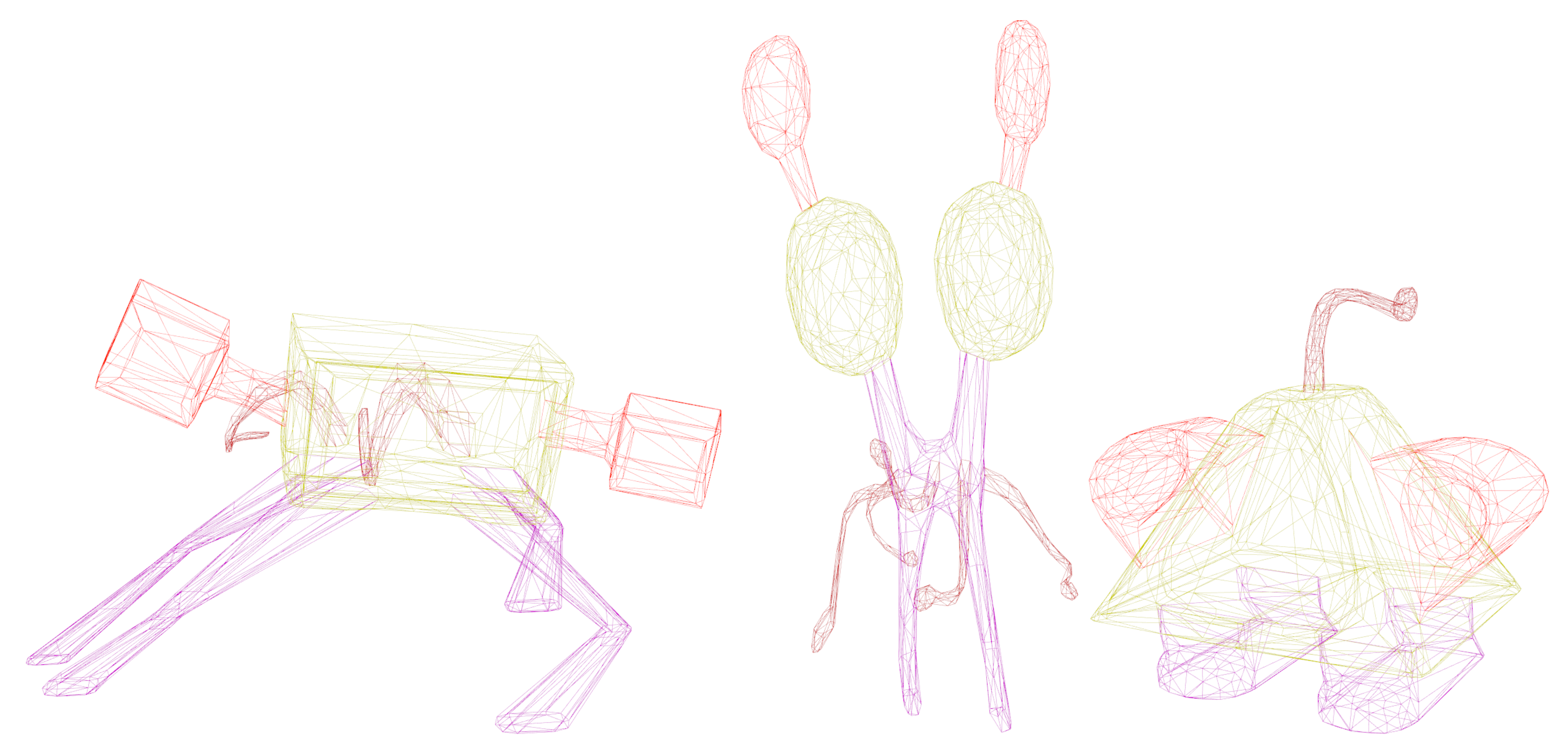}}
		\subfigure[Vases (Ours)]{
			\includegraphics[width=5.5cm]{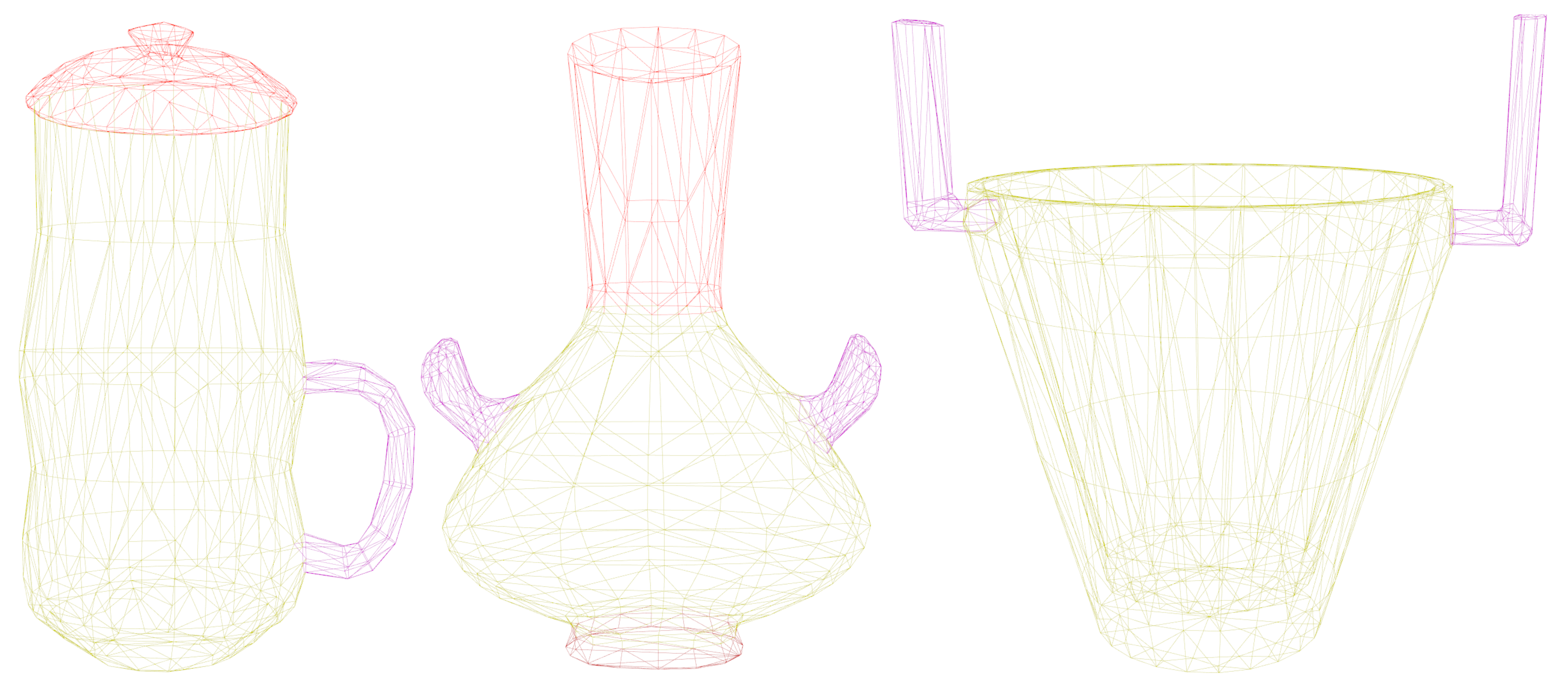}}
		\subfigure[Vases (Ground truth)]{
			\includegraphics[width=5.5cm]{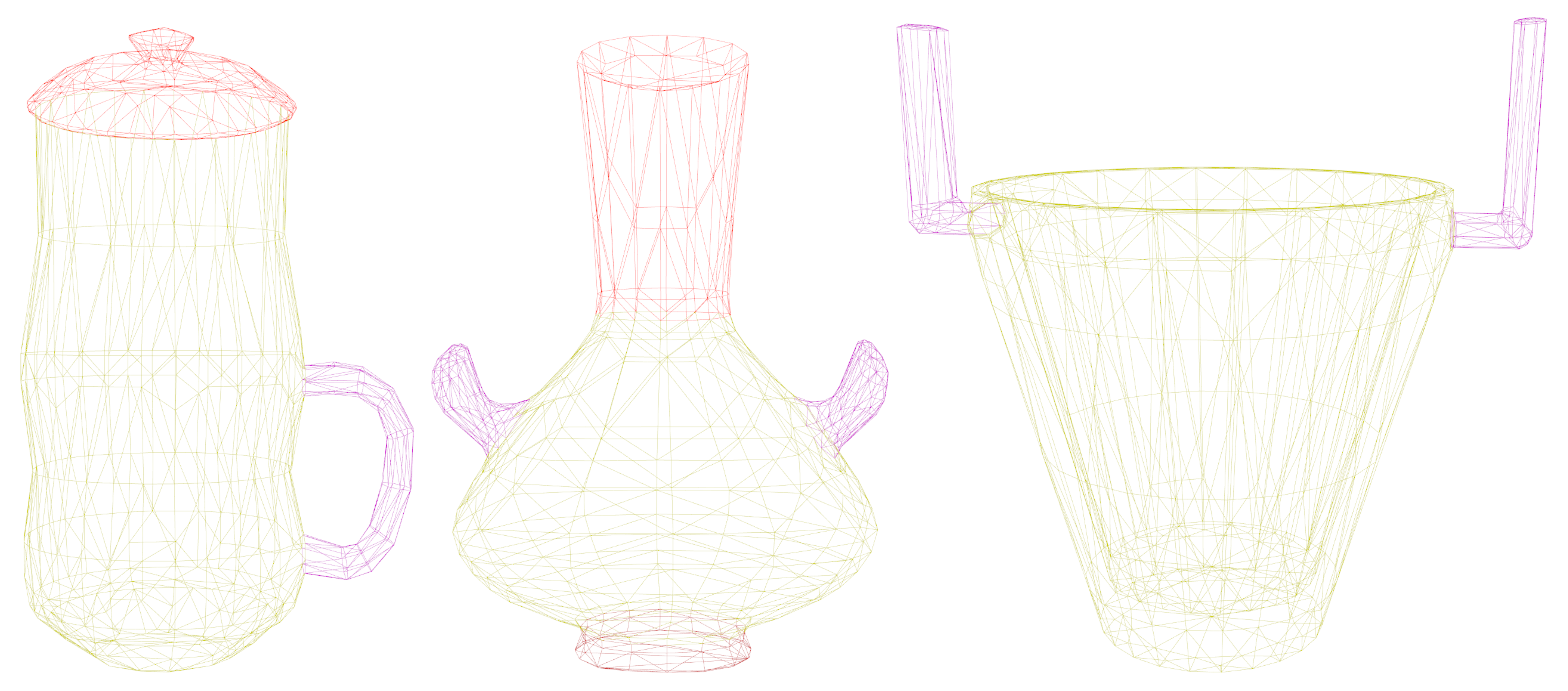}}
		\subfigure[Chairs (Ours)]{
			\includegraphics[width=5.5cm]{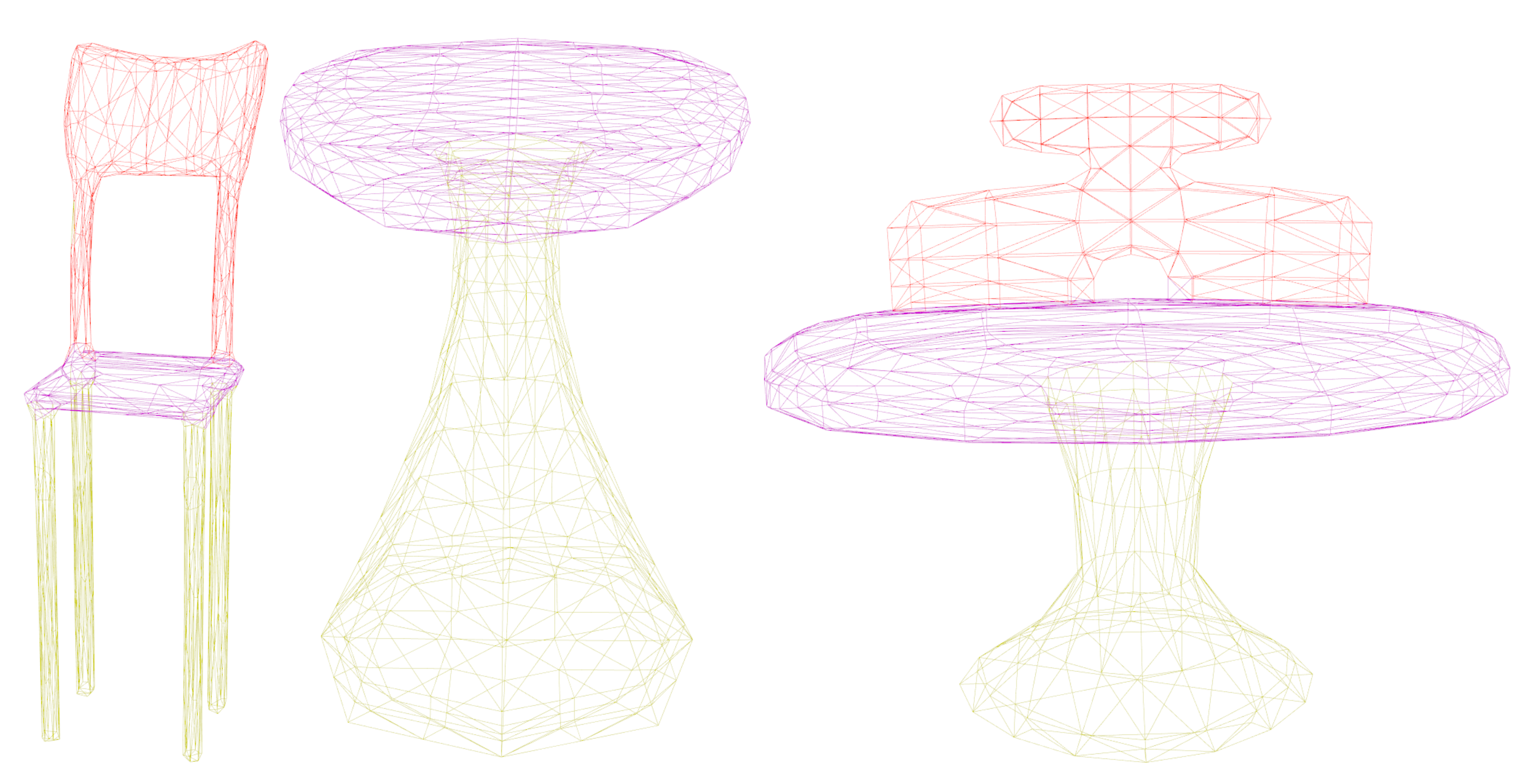}}
		\subfigure[Chairs (Ground truth)]{
			\includegraphics[width=5.5cm]{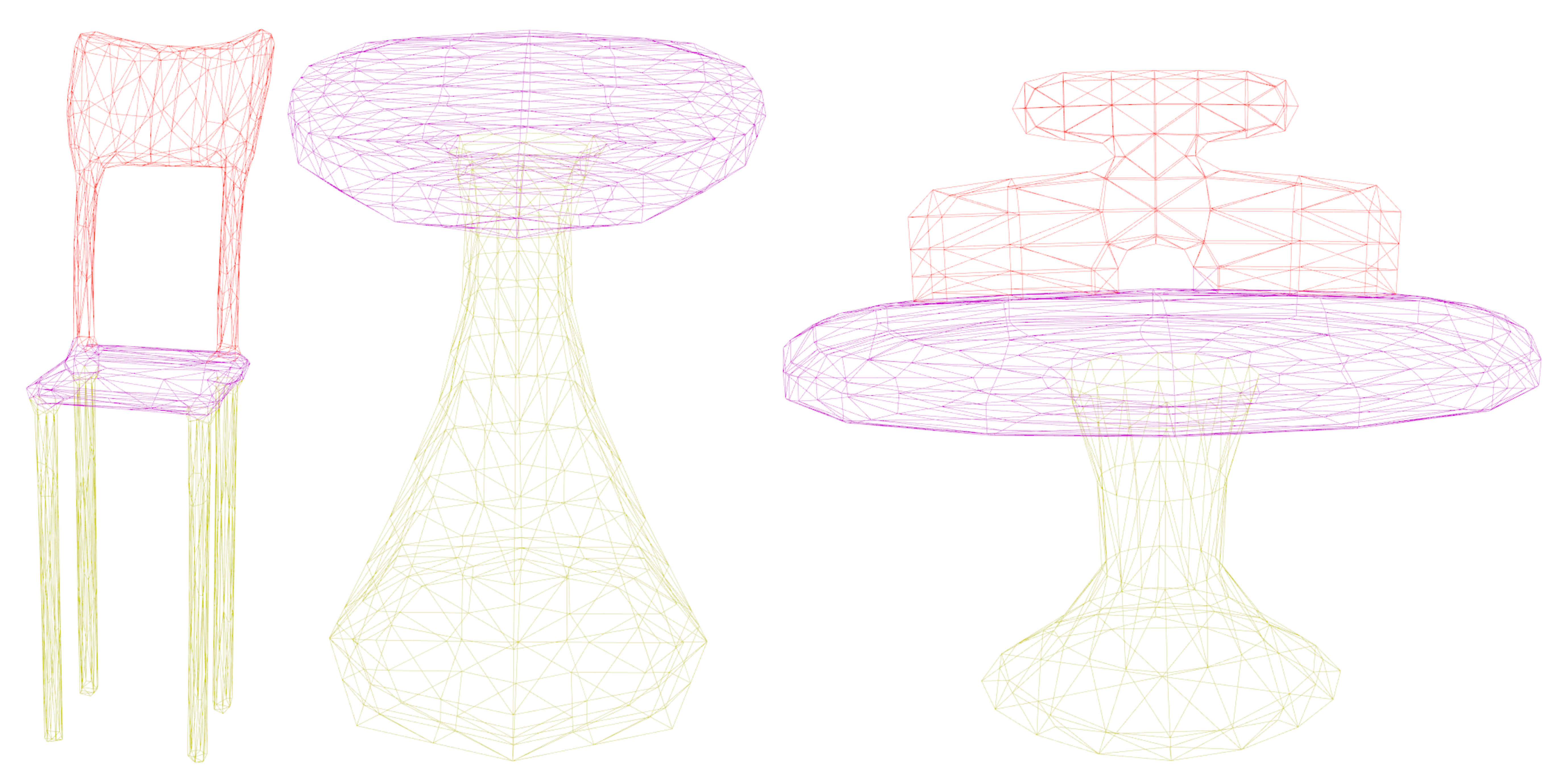}}
		\subfigure[Human Body (Ours)]{
			\includegraphics[width=5.5cm]{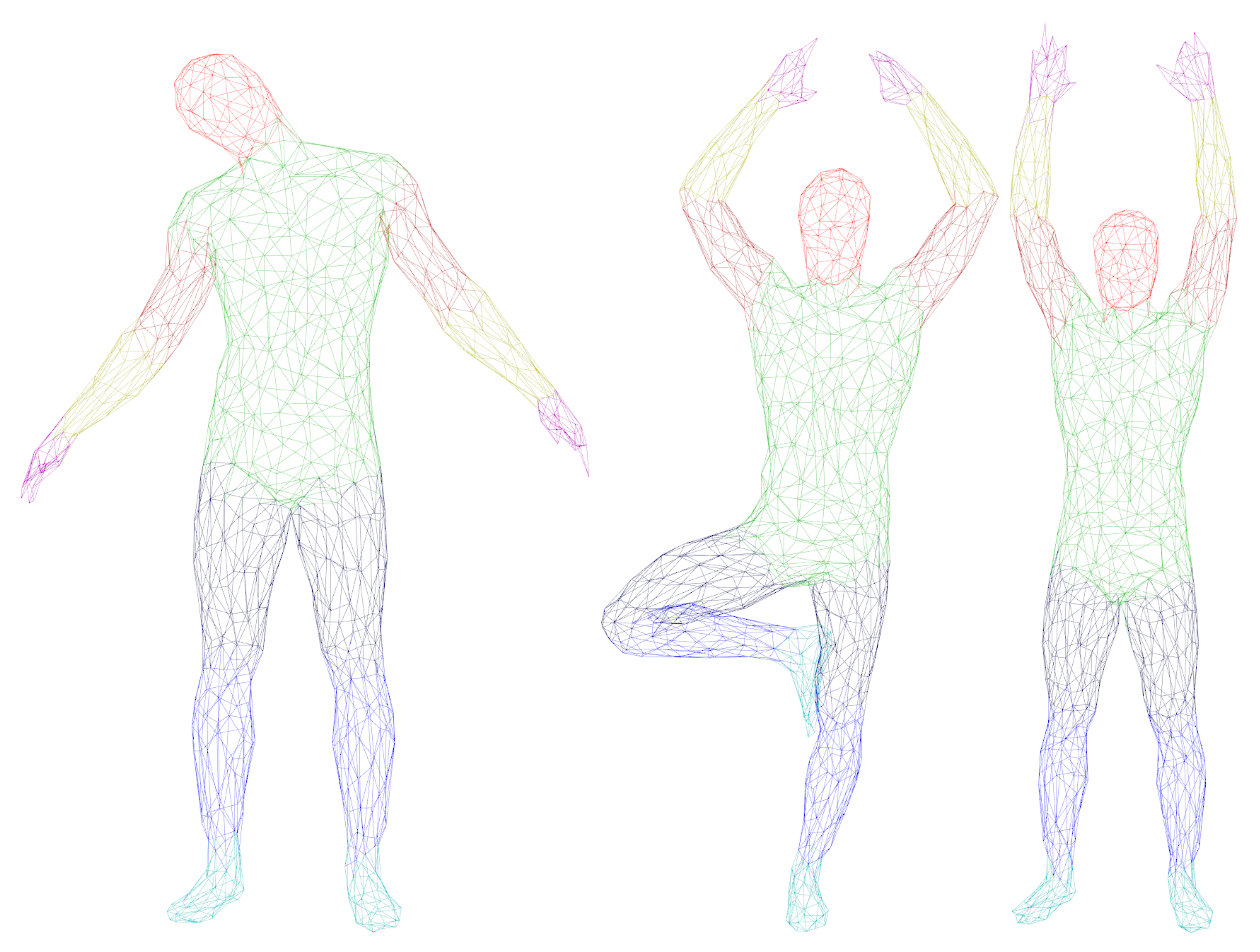}}
		\subfigure[Human Body (Ground truth)]{
			\includegraphics[width=5.5cm]{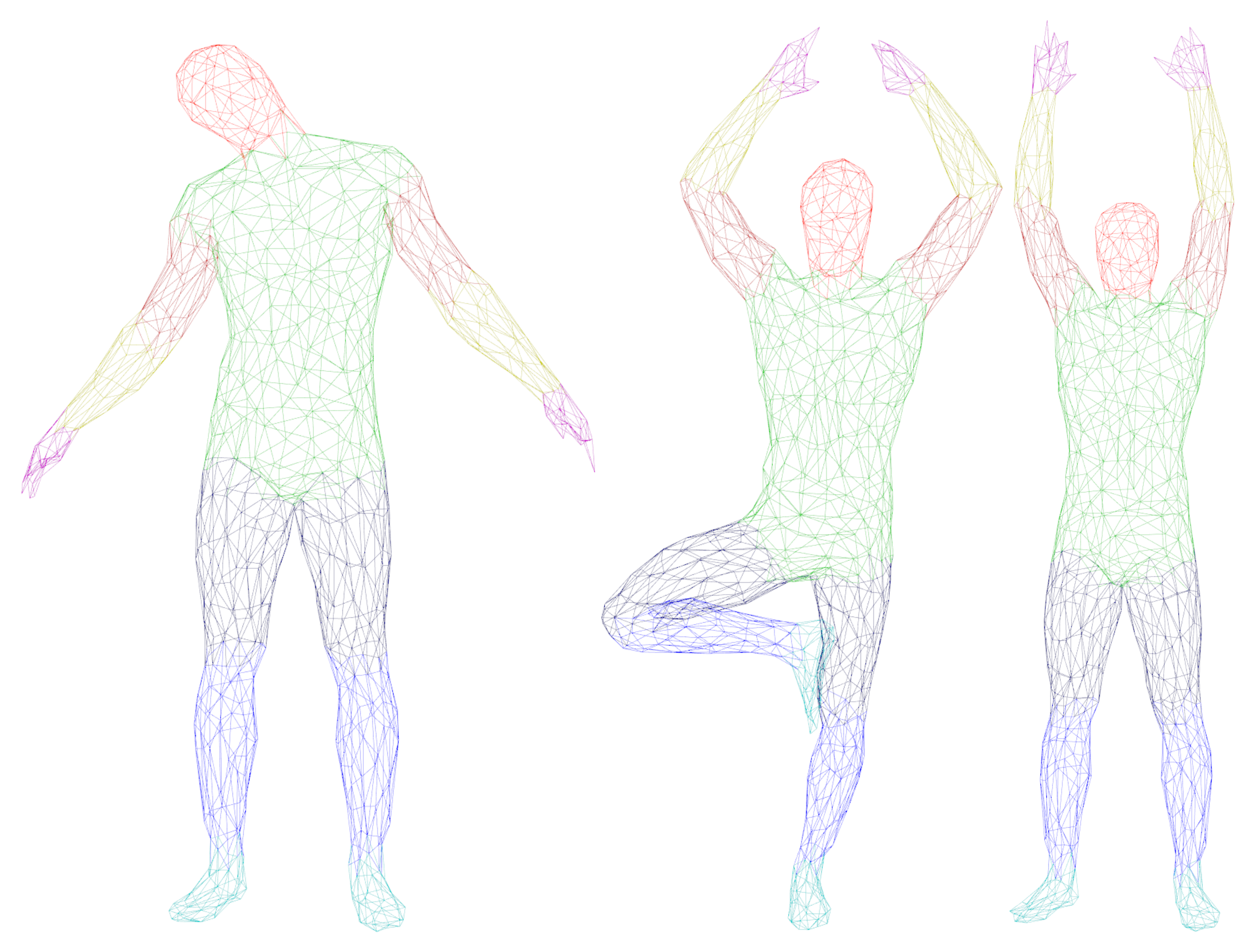}}
		\caption{Visualization of segmentation results with edge labels. (a), (c), (e), (f) are the prediction results of MDC-GCNs on the Telealiens, Vases, Chairs, and Human Body data sets respectively (three randomly selected). (b), (d), (f), (g) are the corresponding ground truth. }
		\label{Fig.edge_labels}
	\end{figure}

	\begin{figure}[htbp]
		\centering
		\subfigure[Telealiens]{
			\includegraphics[width=4.0cm]{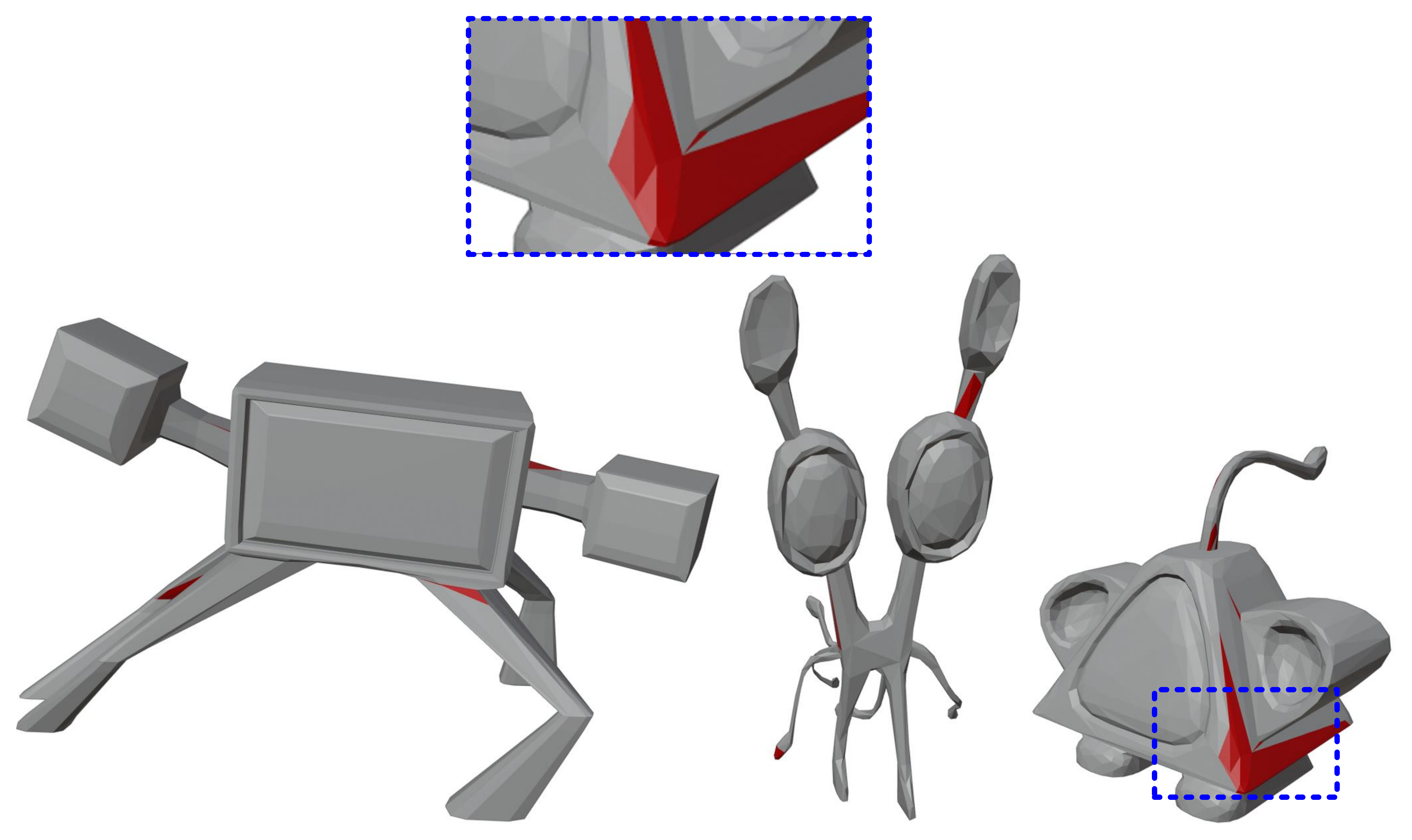}}
		\subfigure[Vases]{
			\includegraphics[width=4.0cm]{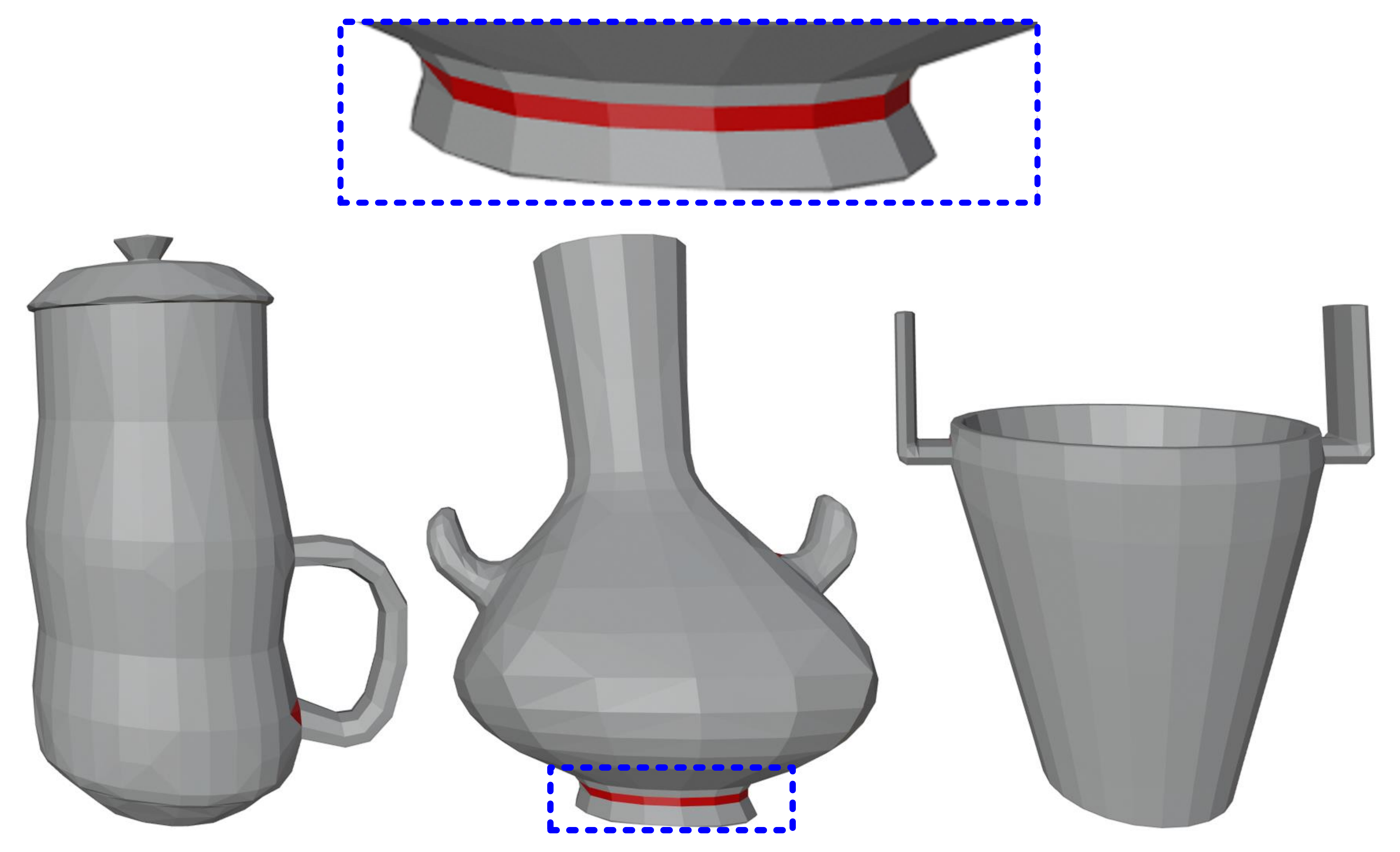}}
		\subfigure[Chairs]{
			\includegraphics[width=4.0cm]{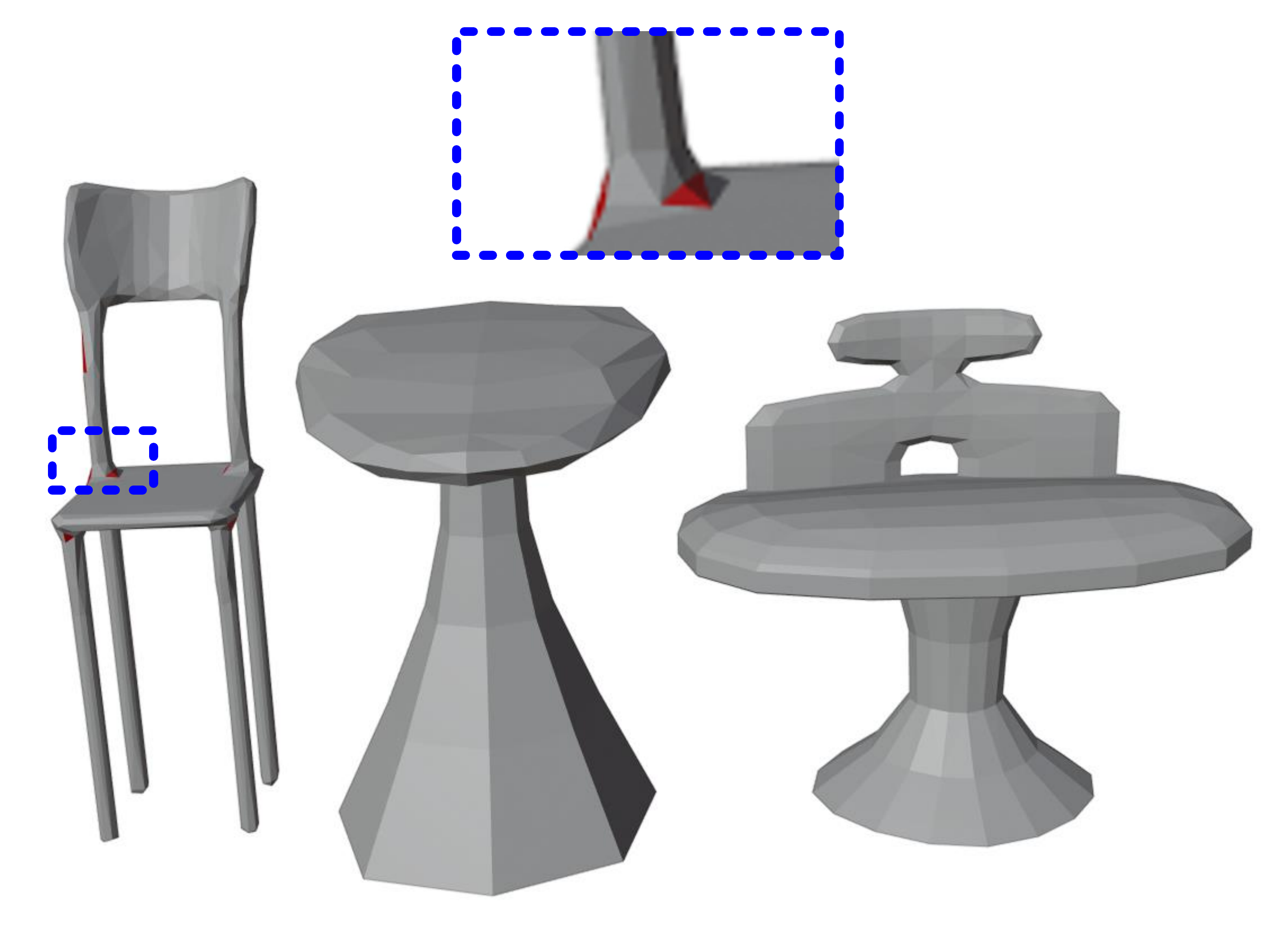}}
		\subfigure[Human Body]{
			\includegraphics[width=4.0cm]{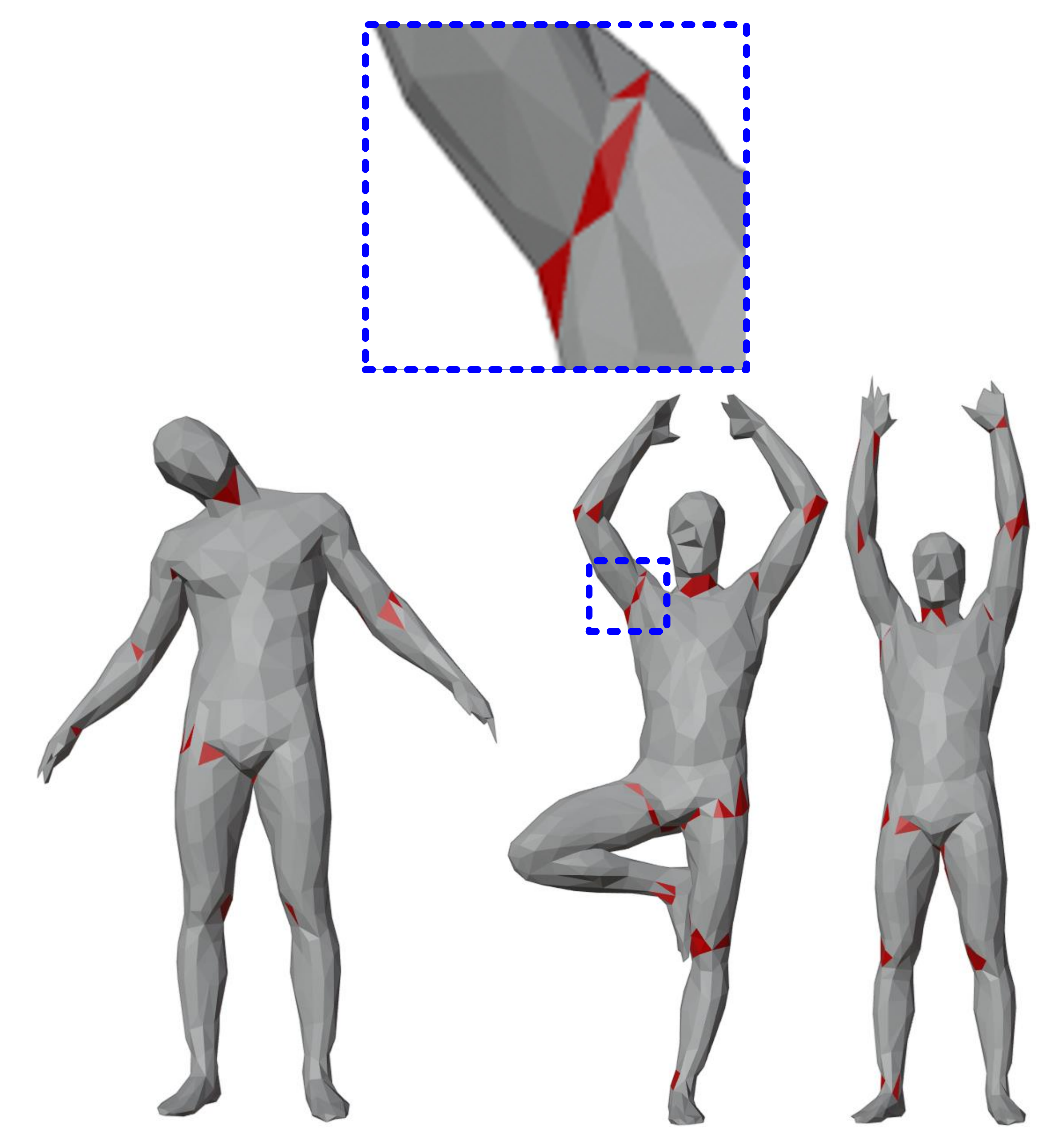}}
		\subfigure[Telealiens]{
			\includegraphics[width=4.0cm]{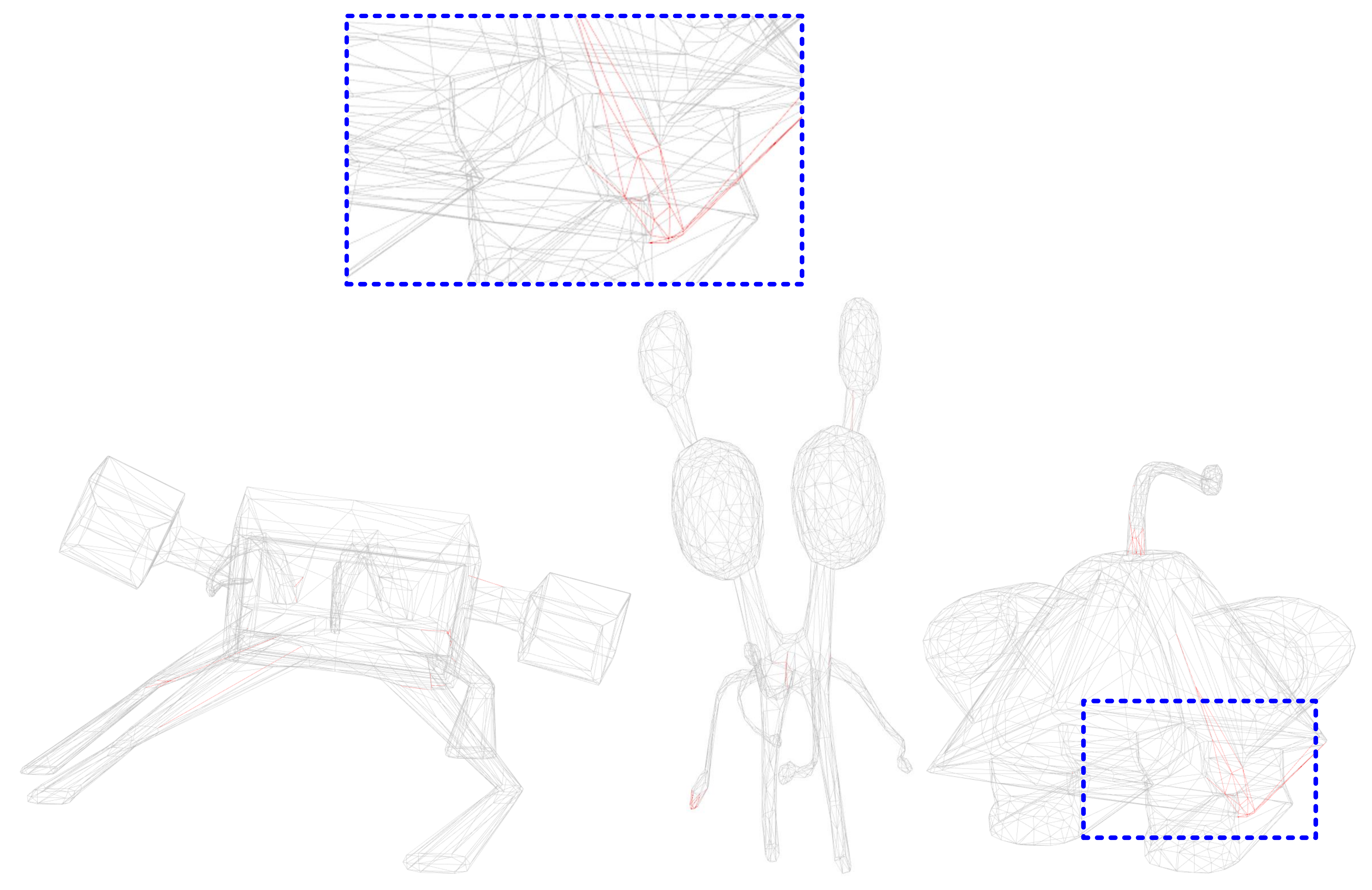}}
		\subfigure[Vases]{
			\includegraphics[width=4.0cm]{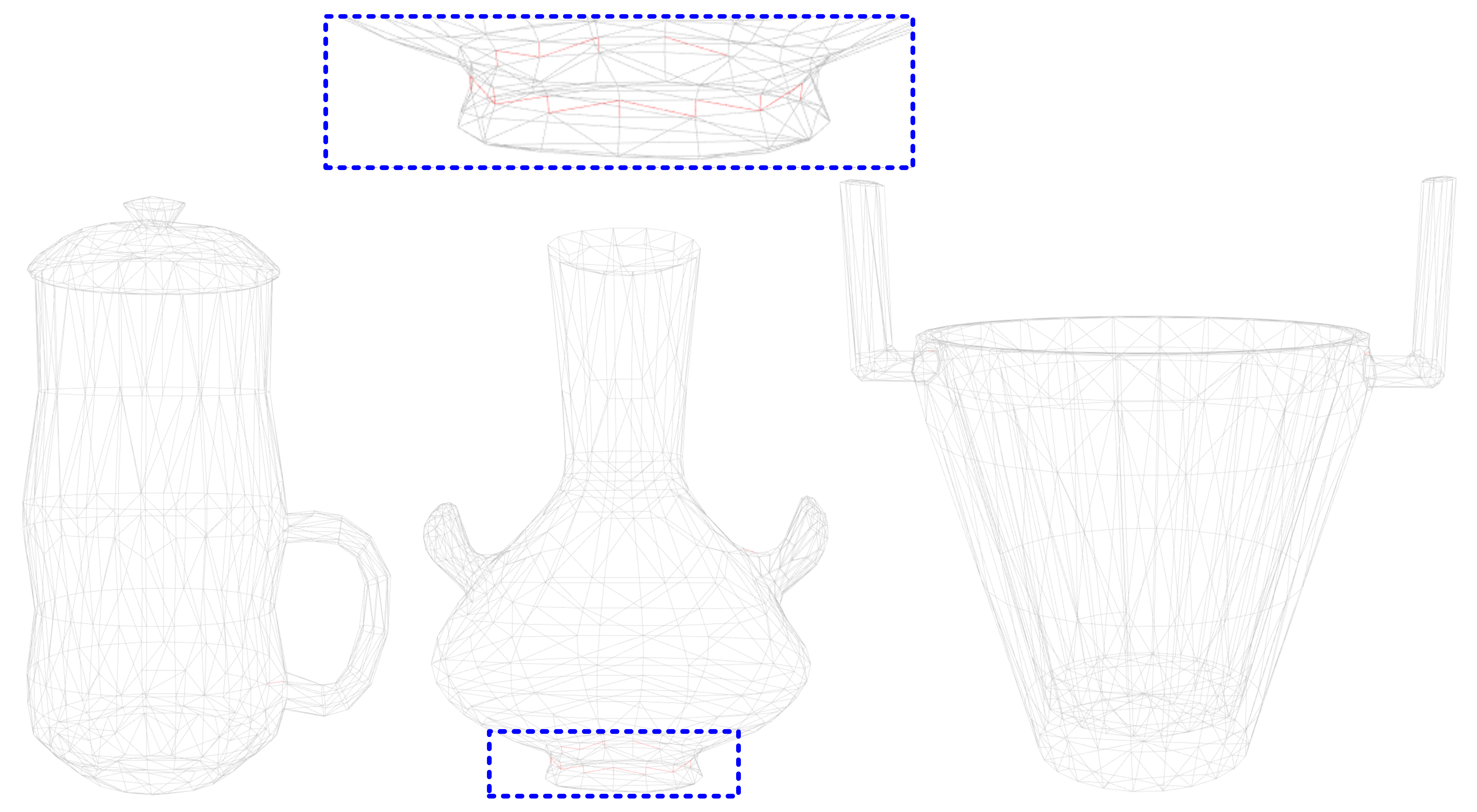}}
		\subfigure[Chairs]{
			\includegraphics[width=4.0cm]{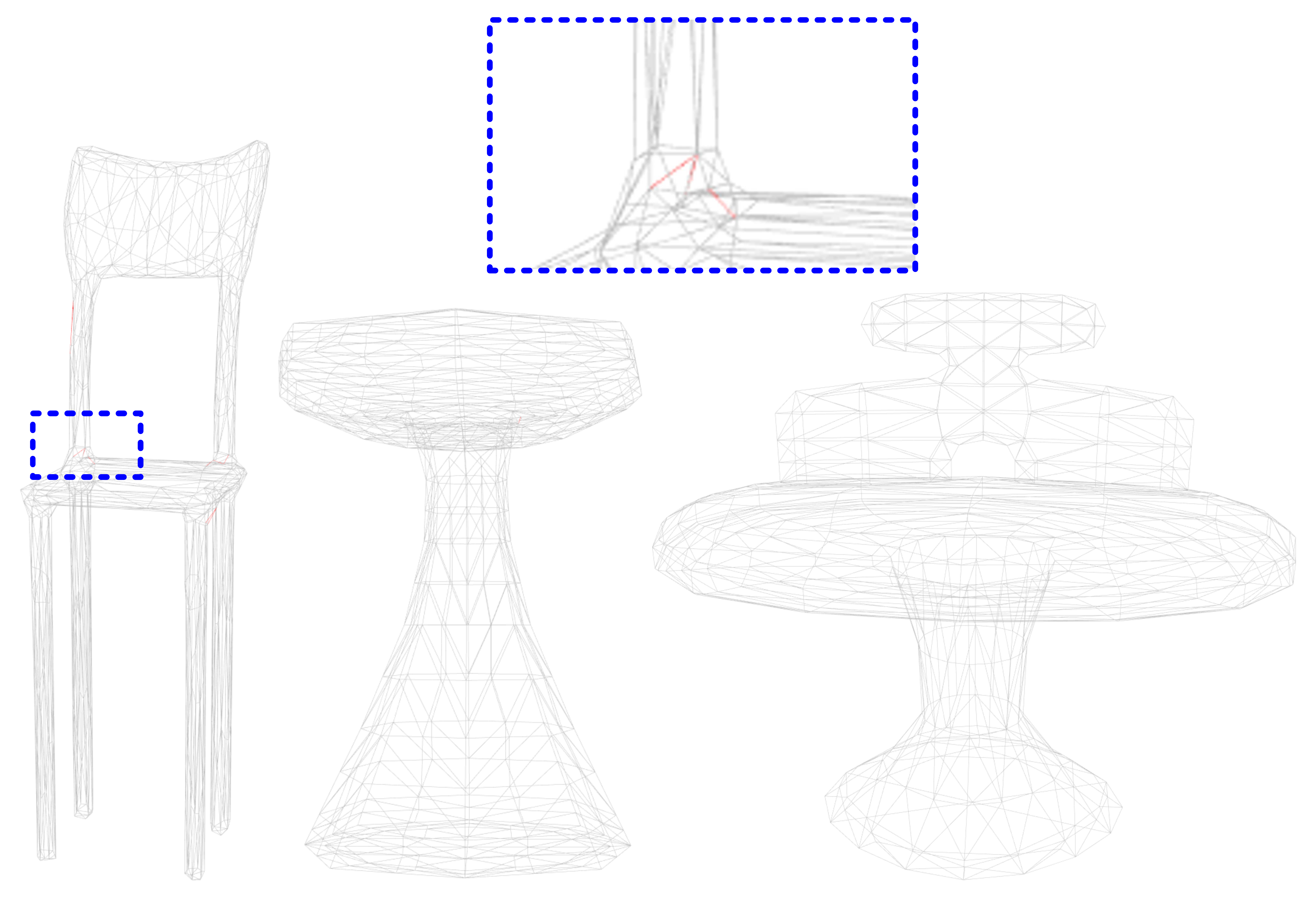}}
		\subfigure[Human Body]{
			\includegraphics[width=4.0cm]{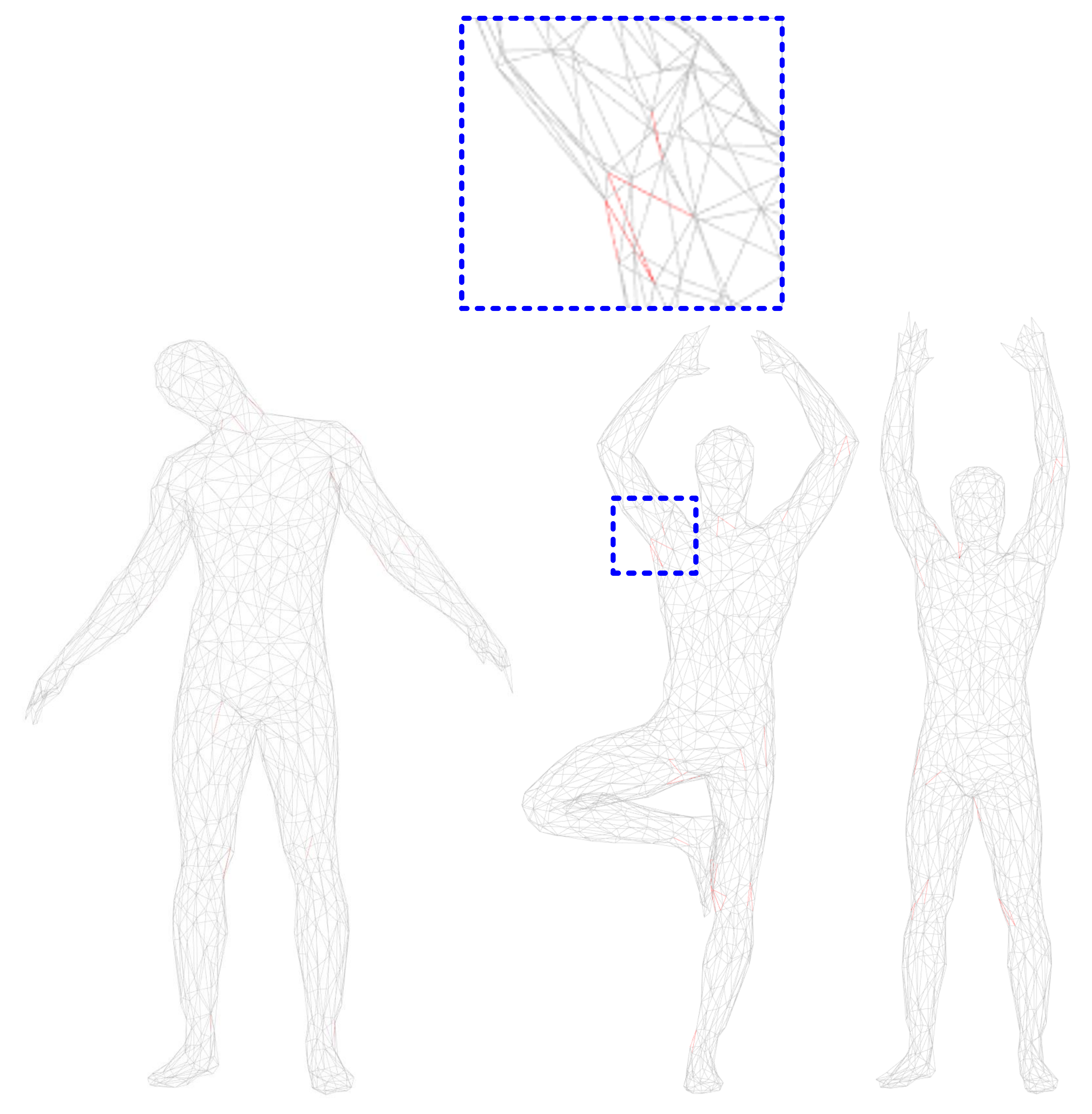}}
		\caption{Visualization of the \textcolor{red}{difference} between segmentation result and ground truth. (a), (b), (c), (d) are the difference between predicted results and ground truth (face labels), (e), (f), (g), (h) are the corresponding edge labels. Most errors made by the MDC-CGNs are located around the junctions between two difference surfaces. Please see the zoomed detail.}
		\label{Fig.difference_labels}
	\end{figure}

	\begin{table}[!]  
		\centering  
		\caption{Classification results of the input graph node features ablation experiment on SHREC and Cube Engraving}  
		\label{fig:Ablation SHREC Classification_results}
		\resizebox{1.0\textwidth}{!}{
			\begin{tabular}{ccccc}  
				\hline
				Feature & Dimensions& SHREC (16) Acc& SHREC (10) Acc&cubes Acc\\ [0.5ex]  
				\hline
				$\{\theta\}$ & 3 & 97.5$\%$& 95.0$\%$  &72.53$\%$ \\
				$\{GC\}$ & 6  & 96.66$\%$& 95.0$\%$& 92.87$\%$ \\
				$\{N_f\}$ & 12 & 93.33$\%$& 78.33$\%$& 85.34$\%$ \\
				$\{P\}$ & 18  & 66.66$\%$& 82.5$\%$& 49.92$\%$ \\
				$\{N_v\}$ & 18  & 95.83$\%$& 89.16$\%$& 86.95$\%$ \\
				${\{N_v\},~\{GC\},~\{N_f\},~\{\theta\}\}}$ & 39 & 98.3$\%$ & 97.5$\%$& 91.81$\%$ \\
				${\{\{P\},~\{GC\},~\{N_f\},~\{\theta\}\}}$ & 39  & 98.3$\%$ & 96.67$\%$& 94.70$\%$\\
				${\{\{P\},~\{N_v\},~\{GC\},~\{\theta\}\}}$ & 45  & 97.5$\%$& 97.5$\%$& 94.08$\%$ \\
				${\{\{P\},~\{N_v\},~\{N_f\},~\{\theta\}\}}$ & 51  & 98.33$\%$& 95.83$\%$& 92.87$\%$ \\
				${\{\{P\},~\{N_v\},~\{GC\},~\{N_f\}\}}$ & 54  & 99.16$\%$& 95.83$\%$& 94.54$\%$ \\
				${\{\{P\},~\{N_v\},~\{GC\},~\{N_f\},~\{\theta\}\}}$ & 57  &\textbf{ 99.7}$\%$ & \textbf{99.2}$\%$& \textbf{94.99}$\%$ \\
				\hline
			\end{tabular}
		}
	\end{table}
	
	\begin{table}[!]  
		\centering  
		\caption{Classification results of hidden layer output graph node features ablation experiment on SHREC and Cube Engraving}  
		\label{fig:Ablation of nodes features SHREC Classification_results}
		\resizebox{1.0\textwidth}{!}{
			\begin{tabular}{ccccc}  
				\hline
				Feature & Nodes& SHREC (16) Acc& SHREC (10) Acc&cubes Acc\\ [0.5ex]  
				\hline
				${\{\{P\},~\{N_v\},~\{GC\},~\{N_f\},~\{\theta\}\}}$ & 8 & 64.16$\%$& 59.16$\%$  &70.10$\%$ \\
				${\{\{P\},~\{N_v\},~\{GC\},~\{N_f\},~\{\theta\}\}}$ & 16  & 78.33$\%$& 74.16$\%$& 82.24$\%$ \\
				${\{\{P\},~\{N_v\},~\{GC\},~\{N_f\},~\{\theta\}\}}$ & 32 & 92.5$\%$& 76.67$\%$& 89.2$\%$ \\
				${\{\{P\},~\{N_v\},~\{GC\},~\{N_f\},~\{\theta\}\}}$ & 64  & 94.1$\%$& 94.1$\%$& 92.71$\%$ \\
				${\{\{P\},~\{N_v\},~\{GC\},~\{N_f\},~\{\theta\}\}}$ & 128  & 96.6$\%$& 95.0$\%$& 93.32$\%$ \\
				${\{\{P\},~\{N_v\},~\{GC\},~\{N_f\},~\{\theta\}\}}$ & 256 & 98.3$\%$ & 93.33$\%$& 93.93$\%$ \\
				${\{\{P\},~\{N_v\},~\{GC\},~\{N_f\},~\{\theta\}\}}$ & 512  & 97.5$\%$ & 96.67$\%$& 94.08$\%$\\
				${\{\{P\},~\{N_v\},~\{GC\},~\{N_f\},~\{\theta\}\}}$& 1024   &\textbf{ 99.7}$\%$ & \textbf{99.2}$\%$& \textbf{94.99}$\%$\\
				\hline
			\end{tabular}
		}
	\end{table}

	\begin{table}[!]\small  
		\centering  
		\caption{Segmentation results of the input  graph node features ablation experiment on COSEG and Human Body}  
		\label{fig:Ablation COSEG_results}
		\resizebox{1.0\textwidth}{!}{
			\begin{tabular}{cccccc}  
				\hline
				Feature & Dimensions & Vases Acc& Chairs Acc& Telealiens Acc& Human Acc \\ [0.5ex]  
				\hline
				$\{\theta\}$ & 3 & 83.94$\%$ & 89.09$\%$ & 70.78$\%$   & 67.92$\%$\\
				$\{GC\}$ & 6   & 88.16$\%$ & 91.72$\%$ & 84.24$\%$ & 85.57$\%$ \\
				$\{N_f\}$ & 12 & 93.69$\%$ & 98.97$\%$ & 88.56$\%$  & 90.92$\%$\\
				$\{P\}$ & 18 & 95.11$\%$ & 85.60$\%$ & 57.25$\%$  & 68.87$\%$\\
				$\{N_v\}$ & 18  & 93.79$\%$ & 99.30$\%$ & 92.61$\%$ & 92.61$\%$\\
				${\{N_v\},~\{GC\},~\{N_f\},~\{\theta\}\}}$ & 39 & 94.47$\%$ & 99.64$\%$ & 94.67$\%$ & 94.50$\%$ \\
				${\{\{P\},~\{GC\},~\{N_f\},~\{\theta\}\}}$ & 39  & 97.41$\%$ & 99.32$\%$ & 93.91$\%$  & 92.38$\%$\\
				${\{\{P\},~\{N_v\},~\{GC\},~\{\theta\}\}}$ & 45  & 97.38$\%$ & 99.61$\%$ & 95.06$\%$  & 93.87$\%$\\
				${\{\{P\},~\{N_v\},~\{N_f\},~\{\theta\}\}}$ & 51  & 97.15$\%$ & 99.65$\%$ & 93.93$\%$  & 92.42$\%$\\
				${\{\{P\},~\{N_v\},~\{GC\},~\{N_f\}\}}$ & 54 & \textbf{97.49}$\%$ & 99.71$\%$ & 94.95$\%$  & 94.27$\%$\\
				${\{\{P\},~\{N_v\},~\{GC\},~\{N_f\},~\{\theta\}\}}$ & 57 & 97.68$\%$ & \textbf{99.74}$\%$ & \textbf{95.78}$\%$  & \textbf{94.65}$\%$\\
				\hline
			\end{tabular}
		}
	\end{table}

	\begin{table}[!]  
		\centering  
		\caption{Segmentation results of hidden layer output graph nodes ablation experiment on COSEG and Human Body}  
		\label{fig:Segmentation results of graph nodes ablation experiment on COSEG and Human}
		\resizebox{1.0\textwidth}{!}{
			\begin{tabular}{cccccc}  
				\hline
				Feature & Nodes & Vases Acc& Chairs Acc& Telealiens Acc& Human Acc \\ [0.5ex]  
				\hline
				${\{\{P\},~\{N_v\},~\{GC\},~\{N_f\},~\{\theta\}\}}$ & 8 & 90.07$\%$& 96.00$\%$  &88.39$\%$ &82.27$\%$\\
				${\{\{P\},~\{N_v\},~\{GC\},~\{N_f\},~\{\theta\}\}}$ & 16  & 94.33$\%$& 98.70$\%$& 89.65$\%$ &84.03$\%$\\
				${\{\{P\},~\{N_v\},~\{GC\},~\{N_f\},~\{\theta\}\}}$ & 32 & 95.49$\%$& 98.96$\%$& 90.61$\%$ &87.52$\%$\\
				${\{\{P\},~\{N_v\},~\{GC\},~\{N_f\},~\{\theta\}\}}$ & 64  & 96.43$\%$& 98.45$\%$& 91.82$\%$&89.73$\%$ \\
				${\{\{P\},~\{N_v\},~\{GC\},~\{N_f\},~\{\theta\}\}}$ & 128  & 96.80$\%$& 98.59$\%$& 92.88$\%$&91.24$\%$ \\
				${\{\{P\},~\{N_v\},~\{GC\},~\{N_f\},~\{\theta\}\}}$ & 256 & 96.97$\%$ & 98.56$\%$& 93.34$\%$&92.66$\%$ \\
				${\{\{P\},~\{N_v\},~\{GC\},~\{N_f\},~\{\theta\}\}}$ & 512  & 97.12$\%$ & 99.17$\%$& 94.00$\%$&93.54$\%$\\
				${\{\{P\},~\{N_v\},~\{GC\},~\{N_f\},~\{\theta\}\}}$& 1024  & \textbf{ 97.68}$\%$& \textbf{99.74}$\%$& \textbf{95.78}$\%$&\textbf{94.65}$\%$ \\
				\hline
			\end{tabular}
		}
	\end{table}

	\subsection{Segmentation}
	When applying MDC-GCNs to the task of segmentation, there is no need to make major changes to the network, and only needs to cascade two DC-blocks through the middle one-layer graph convolution layer, as shown in Fig.~\ref{fig:MDC_GCN_network_segmentation}. Our network does not need to perform image-like down-sampling and up-sampling operations for the mesh structure, but only extracts features through the DC-block, and then outputs node-level classification through graph convolution (different from classification tasks, which are the entire graph structure classification). We will verify the segmentation performances of MDC-GCNs on the dataset COSEG~\cite{wang2012active} and Human Body Segmentation~\cite{maron2017convolutional}.
	\par The three types of 3D meshes in COSEG are Telealiens, Chairs, and Vases. They contain 200, 300 and 400 models of different shapes, respectively. Each model of these three categories contains different numbers of face (Telealiens: 1500, Chairs: 1000, Vases: 1500). We follow the rule of dividing the training and testing sets of MeshCNN, i.e., $85\%$ training and $15\%$ testing, randomly generate 5 training and testing sets, and report the average results of the 5 runs.
	\par Human Body Segmentation contains 370 human body models used in~\cite{anguelov2005scape},~\cite{vlasic2008articulated}, and~\cite{bogo2014faust} and 18 human body models from SHREC07~\cite{giorgi2007shape}. The humanoid models in the dataset each contain 1500 faces. For the sake of fairness, we also randomly divide 5 times according to the rules in the~\cite{maron2017convolutional}, and take the average value. Since MeshCNN uses edge as the unit, in the segmentation comparison experiment, we convert the face label into a label with edge as the unit for comparison. Examples of the visual results are shown in Fig.~\ref{fig:aliens_segmentation} (using the face as the label of the model) and Fig.~\ref{Fig.edge_labels} (using the edge as the label of the model). Visualization of the errors made by the MDC-GCNs are shown in Fig.~\ref{Fig.difference_labels}. The quantitative results of COSEG and Human Body segmentation are shown in Table~\ref{fig:COSEG_results} and~\ref{fig:Human Body Segmentation_results}. It should be noted that the quantitative results of all segmentation comparison experiments use edge as the label.

	
	\section{Ablation studies}
	In order to better verify the 3D shape learning ability of MDC-GCNs, we set up multiple sets of ablation experiments. The ablation experiment included the influence of local feature of the 3D mesh and the dimension of hidden layer output graph nodes on MDC-GCNs.
	\par Table~\ref{fig:Ablation SHREC Classification_results} and~\ref{fig:Ablation of nodes features SHREC Classification_results} respectively correspond to the effects of local features and hidden layer output graph nodes on the classification performance of MDC-GCNs. Table~\ref{fig:Ablation COSEG_results} and~\ref{fig:Segmentation results of graph nodes ablation experiment on COSEG and Human} respectively correspond to the effects of local features and hidden layer output graph nodes on the segmentation performance of MDC-GCNs. All experiments are to maintain the principle of single variable change. From the results of these ablation experiments, the following conclusions can be drawn: 1. It's a better choice to select multi-dimensional geometric features than a single geometric feature for the graph node description.
	2. Structural features such as Gaussian curvature, vertex normals, and surface normals are better than spatial features.
	3. The multi-dimensional geometric feature fusion of spatial and structural features can make the network performance better.
	4. It is better to use 1024 for the graph node dimension when the performance and calculation cost are balanced.
	\par In summary, the choice of 57-dimensional local features and 1024-dimensional graph nodes achieves the best performance.
	
	\section{Conclusion}
	\par In this paper, we propose a novel densely connected graph convolutional networks for 3D meshes. It achieved the classification and segmentation tasks on 3D meshes without data augmentation. We used the face of the mesh as the basic processing unit and introduced the 1-ring face neighbourhood structure to derive multi-dimensional structural features for the graph node to enhance the descriptive power of the graph. We then designed densely connected graph convolutional blocks to aggregate local and non-local features to construct effective and efficient GCNs for 3D object segmentation and classification. Our network is simple and extremely easy to design, and can directly process data in non-Euclidean spaces similar to 3D meshes.
	
	\section{Acknowledgments}

	\section*{References}
	
	\bibliography{mybibfile}
	
\end{document}